%% file: main.tex
\documentclass[manuscript,screen,nonacm]{acmart}

\AtBeginDocument{%
  }

\usepackage{amsmath}
\usepackage{threeparttable}
\usepackage{enumitem}
\usepackage{xspace}
\usepackage{tabularx}
\usepackage{times}
\usepackage{latexsym}
\usepackage{longtable}
\usepackage{float}
\usepackage{multirow}
\usepackage[T1]{fontenc}
\usepackage[utf8]{inputenc}
\usepackage{diagbox}
\usepackage[x11names, table]{xcolor}
\usepackage{makecell}
\usepackage{graphicx}
\usepackage[most]{tcolorbox}
\tcbset{
  colback=gray!10,      
  colframe=gray!40,     
  boxrule=0.5pt,        
  arc=2pt,              
  left=6pt,right=6pt,top=6pt,bottom=6pt
}

\newcommand{\A}{\alpha}
\newcommand{\B}{\beta}
\newcommand{\judge}{judge\xspace}
\newcommand{\judges}{judges\xspace}
\newcommand{\judgess}{judge's\xspace}
\newcommand{\peer}{peer\xspace}
\newcommand{\peers}{peers\xspace}
\newcommand{\peerss}{peer's\xspace}
\definecolor{lightgray}{gray}{0.75}

\begin{document}

\title{Peer Influence across Heterogeneous AI Models}

\author{Frida Nøhr Laustsen}
\affiliation{%
  \institution{IT University of Copenhagen}
  \city{Copenhagen}
  \country{Denmark}}

\author{Marie Haahr Petersen}
\affiliation{%
  \institution{IT University of Copenhagen}
  \city{Copenhagen}
  \country{Denmark}}

\author{Victoria Popa}
\orcid{0009-0007-0862-1820}
\affiliation{%
  \institution{University of Pisa}
  \city{Pisa}
  \country{Italy}}

\author{Ariel Flint}
\orcid{0009-0004-4356-5949}
\affiliation{%
  \institution{City St. George's, University of London}
  \city{London}
  \country{United Kingdom}}

\author{Romualdo Pastor-Satorras}
\orcid{0000-0002-4051-6007}
\affiliation{%
  \institution{Universitat Polit\`ecnica de Catalunya}
  \city{Barcelona}
  \country{Spain}}
  
\author{Andrea Baronchelli}
\orcid{0000-0002-0255-0829}
\affiliation{%
  \institution{City St. George's, University of London}
  \city{London}
  \country{United Kingdom}}

\author{Luca Maria Aiello}
\orcid{0000-0002-0654-2527}
\correspondingauthor
\affiliation{%
  \institution{IT University of Copenhagen}
  \city{Copenhagen}
  \country{Denmark}}
\email{luai@itu.dk}

\renewcommand{\shortauthors}{Nøhr Laustsen et al.}

\begin{abstract}
When two AI agents disagree, who persuades whom? As multi-agent systems increasingly combine language models of different families and sizes, the answer can determine which judgments survive interaction. Measuring persuasion as the probabilistic shift in an agent's decision after a single exchange with a dissenting peer, we test seven open-weight models across three language understanding tasks. We find that persuasion is strong: when models disagree, receivers often abandon their initial judgment after seeing a peer's answer and explanation. Surprisingly, however, neither standalone certainty nor model scale reliably predicts persuasion dynamics. Models producing almost perfectly consistent decisions in isolation can be among the most susceptible to persuasion, and small models can match larger ones as persuaders and resist their influence just as effectively. Furthermore, we show that the size of the shift depends more on the susceptibility of the listener than on the persuasiveness of the speaker. Persuasion patterns are therefore specific to each model pairing, with heterogeneity amplifying persuasion in some combinations and suppressing it in others, allowing a dissenting agent running a small model to overturn the judgments of a much larger one. These findings show that the behavior of interacting models cannot be inferred from their individual properties but must be evaluated in the combinations in which they will operate.
\end{abstract}

\maketitle

\section{Introduction}

Automated decision-making workflows are increasingly relying on decentralized multi-agent systems (MAS) of autonomous AI agents.
Such systems self-organize to handle tasks collaboratively in ways that go well-beyond simple ensembling of models: populations of agents powered by Large Language Models (LLMs) execute subtasks locally and exchange their outputs for further processing or recombination, according to the adaptive topology of the communication network linking them~\cite{vezhnevets2023generative, yang2023auto, wu2023autogen, shen2023hugginggpt, yang2026agentnet}.
This new generation of systems is powering complex workflows across a widening range of domains, including software engineering, healthcare, scientific discovery, and the Internet of Things~\cite{llm-based-multi-agents-for-SE, ai-radiology, Gottweis2026CoScientist,wang2025internet, survey-mas}.

The core mechanism underlying decentralized MAS is the dynamics of agent conditioning through message passing. In its most essential formulation, a \emph{\judge} agent $\A$ must generate an output from a \emph{memory state} consisting of two components: the local context $C_\A$ that the \judge holds about the task, and the output $O_\B$ previously produced by a \emph{\peer} agent $\B$. The judge agent $\A$ thus effectively implements a kernel generator function $f_\A(C_\A, O_\B)$, in which the effect of $O_\B$ on the generation is the \emph{influence} that the \peer agent $\B$ exerts over the \judge. Quantifying this peer influence is essential for a statistical characterization of the outputs that agents generate in interconnected systems. Such characterization is key to estimating the likelihood of emergence of different system-level outcomes, and in turn to assess the stability, vulnerabilities, and alignment of the decision-making system~\cite{de2026conformity, Ashery_2025, flint2026group}.

Surprisingly, however, influence among \peer agents remains underexplored in decision-making MAS. Research that develops and tests these systems has been predominantly concerned with how architectural choices affect benchmark performance of the system as a whole, rather than characterizing microscopic influence dynamics~\cite{park2023generative, yang2026agentnet}. A separate and prolific line of work has examined influence between LLM agents through the lens of opinion dynamics~\cite{chuang2024simulating}, tracking the opinion states of interacting agents~\cite{breum2023persuasivepowerlargelanguage, taubenfeld2024systematic, wang2025decoding, cau2025selective, cau2026network}. These studies have shown that LLM agents may indeed be highly susceptible to peer pressure depending on model type and the direction of the shift~\cite{mehdizadeh2025aiagentsuccumbspeerpressure, zhao2025disagreementsinreasoning, bozdag2026persuadecanframeworkevaluating}, and can exhibit conformity to group norms~\cite{cho2025herd, Cau2025SelectiveAgreementNotSychophancy, song2025llmscanthandlepeer, de2026conformity, bellina2026conformity}. These patterns of influence can create group-level biases that are undetected in isolated models, and ultimately lead to misalignment and performance degradation~\cite{coppolillo2025unmasking, Ashery_2025, wynn2025talkisntcheapunderstanding}. Finally, recent results on agent priming show that behavioral policies of LLM agents may be extremely sensitive to alterations in memory context, suggesting that peer influence may be a nuanced and rich phenomenon~\cite{kumar2026failure}.

However, studies on opinion simulations with LLMs have been limited in two ways. First, they often track shifts in agents’ stances on general opinion topics rather than how agents jointly perform tasks, constraining how readily their findings translate to decision-making systems. Second, they instantiate agents that may differ in assigned personas or communication strategies but are homogeneous in their underlying language model~\cite{yang2026understandingagentscalingllmbased}. Model heterogeneity is an increasingly common condition in MAS, both in open systems, where different players contribute agents of different types to a shared workflow~\cite{price2026moltbook, li2026rise, chen2026openclaw, akkil2026emergenceworld, metr2026huggingface, eigenflux2026}, and in self-adaptive systems, where agents can spawn new agents with different characteristics depending on needs~\cite{song2024adaptive, wang2025megaagent, costa2026agentspawn}.

Very few recent studies have experimented with multi-agent systems of heterogeneous language models, either focusing on negotiation debates~\cite{davidson2024evaluating}, or on the evaluations of the performance of a decision-making systems as a whole~\cite{ye2025xmasbuildingmultiagentsystems, davidson2025collaborationgap}, while disregarding the characterization of microscopic dynamics of influence.

We contribute to fill this gap with an evaluation framework to measure peer influence among heterogeneous agents in joint decision-making tasks. To isolate the influence effect in the message-passing component of decentralized MAS, we propose a minimal dyadic setup with one \judge agent and one \peer agent, working over two rounds. In the first round, both agents perform the same task independently. In the second round, the \judge agent repeats the task while considering both its own prior output and the \peerss output from the first round. This setup captures the most essential message-exchange dynamic of this class of decision-making systems~\cite{multi-agent-debate-du}. To measure influence, we introduce a metric that compares the \judge's output across rounds in a way that accounts for both the stochasticity of the generations and the base model bias. This setup allows us to estimate how influence depends on the pairing of model types, relative model size, and baseline confidence of the model when executing the task in isolation. We run our framework with seven models spanning different architectural families and parameter sizes (Llama 8B and 72B, Gemma 4B and 27B, Qwen 7B and 72B, and GPT-OSS 20B) on three binary text classification tasks for a total of $238,682$ text instances.

We find that agents typically exert strong influence on one another, with large shifts observed across all tasks. The reciprocal degree of influence is generally asymmetric, with varying patterns across model families and a general tendency of smaller models to exhibit competitive persuasive power compared to larger models. Nevertheless, the magnitude of the observed shift is determined primarily by the inherent susceptibility of the \judge model more than the influence power of the \peer model, with some models being generally more prone to influence than others, regardless of the type of model of their peers. Surprisingly, we record frequent instances of backfire effects, in which models reinforce the certainty of their own judgment when exposed to an opposing signal. Contrary to prior work suggesting that opinion shifts are more likely when an agent is initially uncertain~\cite{cho2025herd},
we find that some models remain highly susceptible to influence despite expressing high confidence in their initial judgments.
Finally, model heterogeneity has a non-uniform effect on influence dynamics: some combinations of models lead to more pronounced shifts than their respective homogeneous systems, while others lead to the opposite outcome.

Overall, our work helps enrich the toolkit for assessing and enhancing security in multi-agent systems~\cite{hammond2025multi, brinkmann_machine_2023, tsvetkova_new_2024, anthropic2026multiagent} by providing a methodology to test the reciprocal influence of AI agents in heterogeneous decision-making systems, while highlighting patterns that demonstrate the effect of model heterogeneity on the output stability of these systems.

\section{Experimental Setup}

\subsection{Decision-making framework}

In the first decision-making round, a \emph{\judge} agent $\A$ and a \emph{\peer} agent $\B$ are independently prompted to generate a binary classification label $\hat{y}$ for the same text sample $i$, along with a natural language explanation $e$ motivating their decision. Because our interest is in measuring influence, we focus on cases where the labels of the \judge and the \peer diverge (the more trivial case of convergent labels is discussed in Section~\ref{app:agreeing-influence}). In the second round, the \peerss output is passed to the \judge, which is asked to reconsider its round-1 decision in light of the opposing label and explanation. The two agents can be instantiated from any combination of models, supporting the study of both homogeneous and heterogeneous peer interactions, and they can swap roles for a second run of round 2, so that each pair of models acts as both \judge and \peer on exactly the same set of samples.

Comparing the \judge decision in round 1 ($\hat{y}_\A$) with its decision in round 2 after peer conditioning ($\hat{y}_{\A|\B}$) reveals the influence of the \peer on the \judgess decision-making. We make this comparison probabilistic by repeating each round $M$ times and measuring how much the probability of the \judge generating label $y$ in isolation ($p^{y}_\A$) shifts after the interaction ($p^{y}_{\A|\B}$). The resulting \emph{influence score} $\Delta_{\A|\B} \in [0,1]$ quantifies the influence of the \peer over the \judge across all samples, with $\Delta=0$ denoting no effect and $\Delta=1$ a maximal probability shift. We also compute a complementary \emph{backfiring score} $\Delta^-_{\A|\B} \in [-1,0]$, which captures the extent to which the \judge instead reinforces its prior label preference after exposure to the \peer opinion. To make influence scores comparable across all model pairs, we account for stochasticity in agent generations, individual model biases towards specific labels, and differences in the empirical availability of label combinations (details in Methods).

We interpret the influence results in light of three model-level properties: i) \emph{persuasiveness}, the average influence score obtained when acting as a \peer; ii) \emph{susceptibility}, the average influence score obtained when acting as a \judge; and iii) classification \emph{certainty}, the fraction of samples that the model classified consistently with the same label across all $M$ repetitions in round 1 (i.e., $\max(p^{y=0}_\A, p^{y=1}_\A)$).

\subsection{Datasets}

We experiment with three binary classification datasets in the domain of natural language understanding, spanning different levels of difficulty. The binary labels are reasonably balanced in all datasets, in the $50\%$--$54\%$ range. Data sampling and preprocessing are detailed in Section~\ref{app:preprocessing}.

The first dataset is the \emph{Sentiment Analysis} corpus from the SocKet repository~\cite{DBLP:journals/corr/KhodakSV17_sarcasmDS}, consisting of $91{,}964$ statements paired with human-assigned \emph{Positive}/\emph{Negative} sentiment labels. Assessing sentiment from text is a relatively easy language understanding task, as the text often carries explicit signals of the underlying sentiment.
The second is CommonsenseQA~$2.0$~\cite{talmor2022commonsenseqa20exposinglimits}, consisting of $9{,}264$ factual questions paired with binary \textit{Yes}/\textit{No} answers. Although the questions are intentionally designed to be challenging, they are unambiguous and can often be answered through common-sense reasoning.
The last is the \emph{Sarcasm Detection} dataset from the SocKet repository~\cite{DBLP:journals/corr/KhodakSV17_sarcasmDS}, consisting of $137{,}454$ claims labeled by human annotators as \emph{Literal}/\emph{Sarcastic}. Because identifying context-dependent sarcasm in short text snippets is difficult even for human annotators, this dataset carries inherently less information useful for classification and therefore represents the most challenging of the three tasks.

\subsection{Models and parametrization}

We evaluate a diverse set of open-weight models spanning different parameter sizes, vendors, and geographical regions of development: Google's \texttt{gemma-3-4b} and \texttt{gemma-3-27b}; Meta's \texttt{llama-3.1-8b} and \texttt{llama-3.3-70b}; Alibaba's \texttt{qwen-2.5-7b} and \texttt{qwen-2.5-72b}; and OpenAI's \texttt{gpt-oss-20b}. For brevity, we omit version identifiers and refer to the models by the \texttt{model-size} naming convention. We use 4-bit \texttt{bitsandbytes} quantized versions of all models.
We consider all pairwise configurations of the largest models from each family and, to test effect of model size, we separately pair large and small models within the same family. 
We adopt default sampling parameters when available, or values reported as suitable by previous work as alternatives (details in Section~\ref{app:model-specifications}). Instruction prompts are kept short and with neutral tone, and are detailed in Section~\ref{app:prompts}. In all experiments, we set the number of repetitions per sample to $M=10$.

\section{Results}

\begin{table*}[t!]
\small
\centering
\begin{tabular}{l ccc ccc ccc}
\toprule
\multirow{2}{*}{\textsc{Model}}
& \multicolumn{3}{c}{\textsc{Sarcasm}}
& \multicolumn{3}{c}{\textsc{Commonsense}}
& \multicolumn{3}{c}{\textsc{Sentiment}} \\
\cmidrule(lr){2-4} \cmidrule(lr){5-7} \cmidrule(lr){8-10}
 & \small $+$ & \small $-$ & \small $+$$/$$-$
& \small $+$ & \small $-$ & \small $+$$/$$-$
& \small $+$ & \small $-$ & \small $+$$/$$-$ \\
\midrule
    \texttt{gemma-27b}  & .916 & .063 & .021 & .479 & .495 & .026 & .462 & .524 & .014 \\
    \texttt{gemma-4b}   & .959 & .033 & .008 & .367 & .580 & .053 & .424 & .563 & .013 \\
    \texttt{llama-70b}  & .733 & .192 & .075 & .430 & .536 & .034 & .510 & .475 & .015 \\
    \texttt{llama-8b}   & .855 & .003 & .142 & .433 & .132 & .435 & .482 & .429 & .089 \\
    \texttt{qwen-72b}   & .821 & .128 & .051 & .389 & .563 & .048 & .456 & .527 & .017 \\
    \texttt{qwen-7b}    & .982 & .014 & .004 & .193 & .776 & .031 & .482 & .499 & .019 \\
    \texttt{gpt-20b}    & .188 & .231 & .581 & .191 & .451 & .358 & .376 & .461 & .163 \\
\bottomrule
\end{tabular}
\caption{\textbf{Summary of classification certainty in round 1}. Fraction of samples that, over $M$ repetitions, are assigned always a positive label ($+$), always a negative label ($-$), or a mixed set of labels ($+$$/$$-$). Values are presented across the seven models and three datasets.}
\label{tab:consistent}
\end{table*}

Table~\ref{tab:consistent} provides an overview of the label distribution in the first classification round, after discarding the very small number of samples that consistently produce invalid output formatting (see Section~\ref{app:discarded-claims}). Classification certainty is generally very high, with most models assigning a single label to each sample across all $M$ repetitions. The notable exceptions are \texttt{gpt-20b}, which labels between $16\%$ and $58\%$ of samples with both labels depending on the task, and \texttt{llama-8b}, which labels $43\%$ of samples inconsistently in the Commonsense dataset. Among the samples with full label certainty, the distribution over the two classes varies with task difficulty, despite the balance of the underlying ground truth. It ranges from relatively balanced in the Sentiment dataset to strongly skewed toward the positive label in the Sarcasm dataset. The weak signal available for inferring sarcasm from text likely causes most agents to collapse their label choice toward the bias of their underlying language model.

\begin{figure*}[h]  \includegraphics[width=1\textwidth]{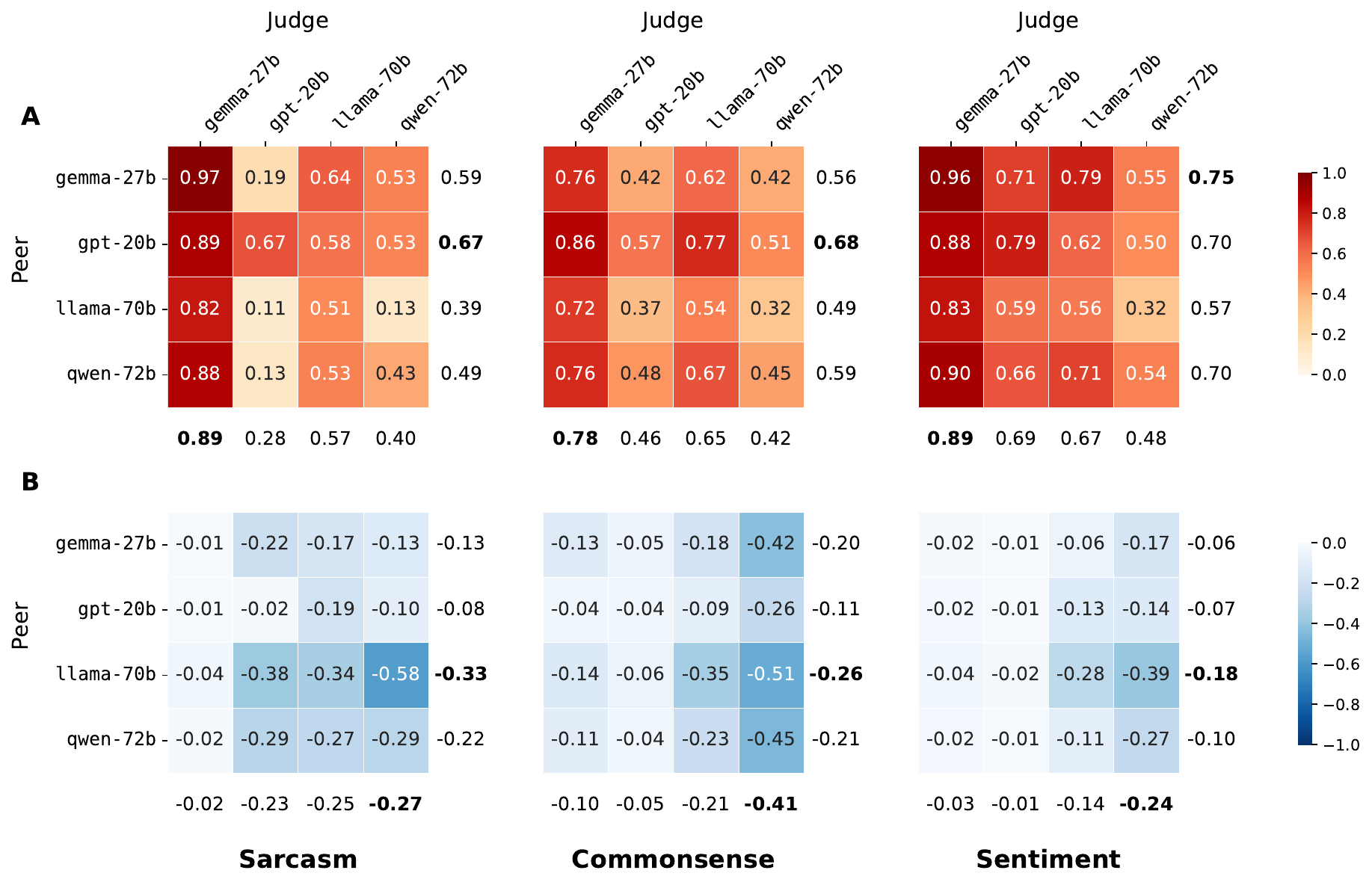}
    \caption{\textbf{Influence scores $\Delta_{\A|\B}$ across model pairs and datasets.} For the positive influence scores (A), column-wise marginal means show how susceptible a \judge is to influence from different \peers, while row-wise marginal means represent the persuasiveness of a \peer across different \judges. For the backfiring scores (B), columns show how strongly a \judge backfires in reaction to peer influence, and rows show how strongly a \peer induces backfiring in \judges, when a backfire event occurs. In both panels, the anti-diagonal reports the outcomes of homogeneous systems, and the off-diagonal entries those of heterogeneous systems. The largest absolute marginals in each row and column are in bold. The margin of error at $95\%$ confidence is $\leq$$0.01$ for all combinations of \judge, \peer, and dataset.}
  \label{fig:large-combined-overall}
\end{figure*}

\subsection{Peer Influence}

Figure~\ref{fig:large-combined-overall}A reports the influence scores $\Delta_{\A|\B}$ for all pairs of the largest models. The marginal means of the matrices represent the \emph{susceptibility} to influence when a model acts as a \judge (columns) and its \emph{persuasiveness} when it acts as a \peer (rows). Figure~\ref{fig:large-combined-overall}B reports the backfiring scores $\Delta^-_{\A|\B}$ instead. The results support five main findings.

First, \peers exert a substantial effect on \judges. Opinion shifts are generally large across models and datasets, with influence scores exceeding $0.5$ in most cases and frequent peaks above $0.75$. 
Regardless of the \judgess label probability distribution in round 1, the probability mass of assigning the \peerss suggested label in round 2 concentrates heavily toward 1 (see Section~\ref{app:influence-heatmaps}).
This finding is in line with prior observations that the judgments of LLM-based agents are typically highly sensitive to external input~\cite{cho2025herd, wu2025can}. 

Second, influence tends to be directional. Within a fixed model pair, an agent that is an influential \peer is likely to be less susceptible to influence when assuming the role of \judge. For example, when pairing \texttt{qwen-72b} with \texttt{llama-70b}, \texttt{qwen-72b} consistently achieves higher influence scores as \peer and lower scores as \judge. At the aggregate model level, however, high persuasiveness is not necessarily associated with low susceptibility: a model may exert strong influence as a \peer while remaining highly susceptible as a \judge. \texttt{Gemma-27b} is the clearest example, exhibiting the highest susceptibility values while also scoring high in persuasiveness.

Third, high certainty does not imply lower susceptibility. The \judges with the highest susceptibility are also those with the highest rate of consistent classification labels when evaluating samples independently. In the Sarcasm dataset, \texttt{gemma-27b} produces consistent labels for $99.95\%$ of samples, yet its susceptibility as a \judge is $0.89$. The pattern holds across the board, as reflected by a positive Spearman rank correlation between certainty (i.e., the proportion of samples labeled consistently in round 1) and susceptibility, when pooling all model pairs and datasets ($\rho=0.61$, $p=0.036$).

Fourth, the final classification outcome is driven more by the characteristics of the \judge than by those of the \peer, as indicated by the higher standard deviation of susceptibility values across all model-dataset combinations ($\sigma=0.19$), compared to persuasiveness values ($\sigma=0.10$).

Fifth, the likelihood that a \peer-\judge interaction backfires is low but not negligible, averaging at about $15\%$ in Sarcasm $12\%$ in Commonsense and $7\%$ in Sentiment, and with peaks exceeding $40\%$ for some combinations of models. The full set of values is reported in Section~\ref{app:backfiring-probability}. Backfiring is also milder overall than positive influence, with peak values rarely falling below $-0.5$.

\vspace{4pt}\noindent\textit{Model overview.} Individual models exhibit distinct influence profiles. \texttt{Gemma-27b} is the most susceptible \judge across all datasets ($0.78$ to $0.89$) and near-zero backfiring throughout. Its shifts are not only frequent but close to complete: post-interaction certainty is pushed near $1.0$ across the full range of baseline probabilities (see Section~\ref{app:influence-heatmaps} for detailed statistics). Despite this, it is also a highly effective \peer, scoring consistently high against every \judge.
\texttt{Gpt-20b} is the only model that combines strong persuasiveness with stability under influence. The contrast is sharpest on Sarcasm, where it reaches influence scores of $0.53$ to $0.89$ as a \peer and the lowest susceptibility of any model ($0.28$). Its backfiring scores are close to zero except on Sarcasm ($-0.23$). \texttt{Llama-70b} behaves similarly, though it is somewhat more susceptible, less persuasive, and backfires more strongly.
\texttt{Qwen-72b} is a more ambiguous case. Judging by positive influence alone, it appears resistant, with marginals below those of \texttt{gemma-27b} and \texttt{llama-70b}. The backfiring scores add nuance to this interpretation: \texttt{qwen-72b} is the \judge with the strongest backfiring effect on all three datasets ($-0.27$, $-0.41$, and $-0.24$ for Sarcasm, Commonsense, and Sentiment). Much of the influence it receives surfaces as movement away from the \peer opinion rather than as stability, most prominently when interacting with \texttt{llama-70b} (details in Section~\ref{app:influence-heatmaps}).

\subsection{Homogeneous vs. heterogeneous interactions} \label{subsec:homo-vs-hetero}

\begin{figure*}[t] 
    \includegraphics[width=1\textwidth]{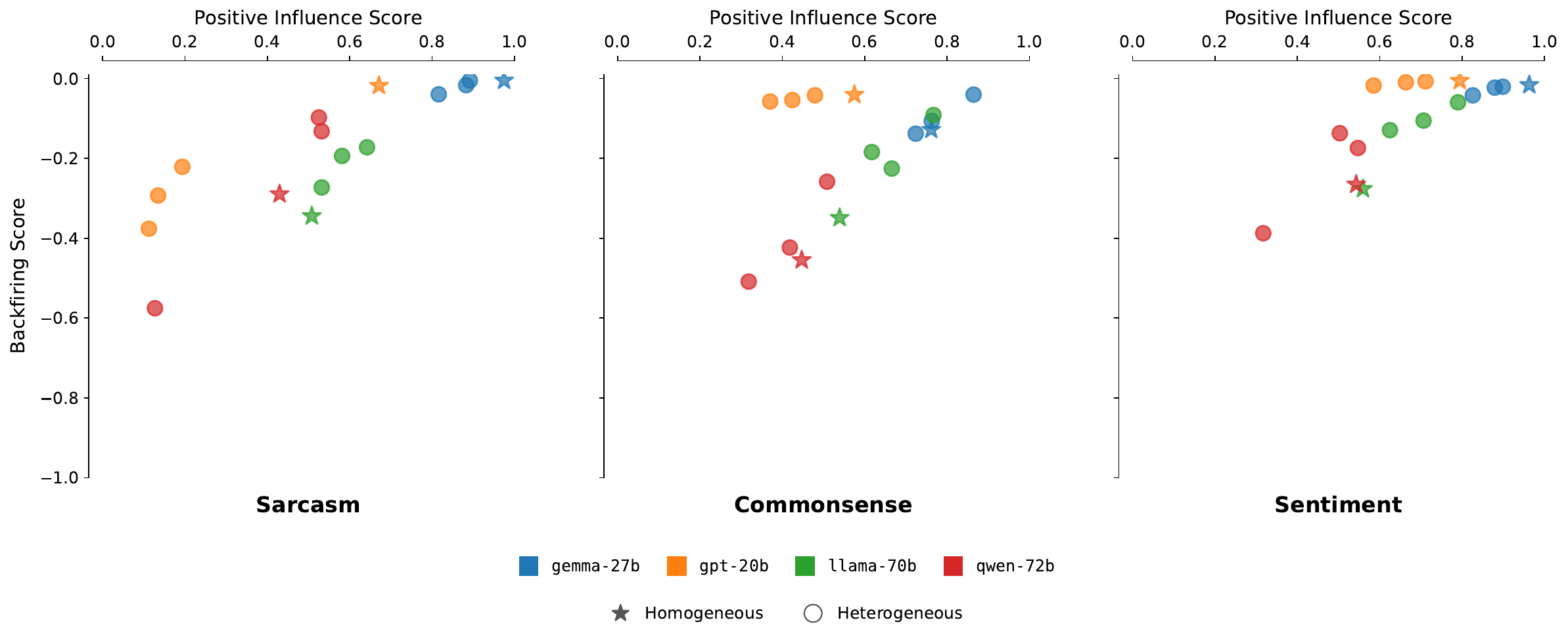}
    \caption{\textbf{Role of model heterogeneity in influence.} Model pairs plotted as a function of their positive influence score (x-axis) and backfiring score (y-axis), colored by \judge agent. Pairs with \judges impervious to influence lie closer to the origin. Stars and circles denote homogeneous and heterogeneous systems, respectively. The distance between stars and circles of the same color indicates behavioral divergence between the two settings.}
    \label{fig:homo-vs-hetero}
\end{figure*}

To compare the decision-making stability of heterogeneous and homogeneous systems, in Figure~\ref{fig:homo-vs-hetero} we plot each model combination as a function of its influence scores ($\Delta_{\A|\B}$) and backfiring scores ($\Delta^-_{\A|\B}$) when acting as \judge.

For any given \judge, the values of all \peers cluster relatively tightly, confirming that the influence outcome is determined mostly by the behavior of the \judge rather than by the \peer. The important role of the \judge over the \peer is further evidenced by the absence of any consistent pattern when the same values are plotted from the \peerss perspective (see Section~\ref{app:homo-vs-hetero}).

Homogeneous configurations often occupy peripheral positions within these clusters, as their outcomes tend to be more distinctive than those of heterogeneous ones, though in ways that depend on the model-dataset combination. The clearest gaps appear for \texttt{gpt-20b} and \texttt{llama-70b}. Moving from a homogeneous to a heterogeneous system, \texttt{gpt-20b} becomes less susceptible, with influence scores dropping by $17\%$ to $79\%$ across datasets, when averaging over all \peer models. \texttt{Llama-70b} moves in the opposite direction, becoming more susceptible in heterogeneous systems by $13\%$ to $27\%$, though simultaneously becoming less likely to experience backfire effects. Detailed percentage differences for all systems are reported in Section~\ref{app:homo-vs-hetero}. The remaining models show less pronounced separation between the two settings.

Overall, these results suggest that heterogeneity does not uniformly raise or lower susceptibility to influence, but instead reshapes influence dynamics in a model-specific way.

\subsection{Impact of model size}

\begin{figure*}[t!] 
    \centering
    \includegraphics[width=0.49\textwidth]{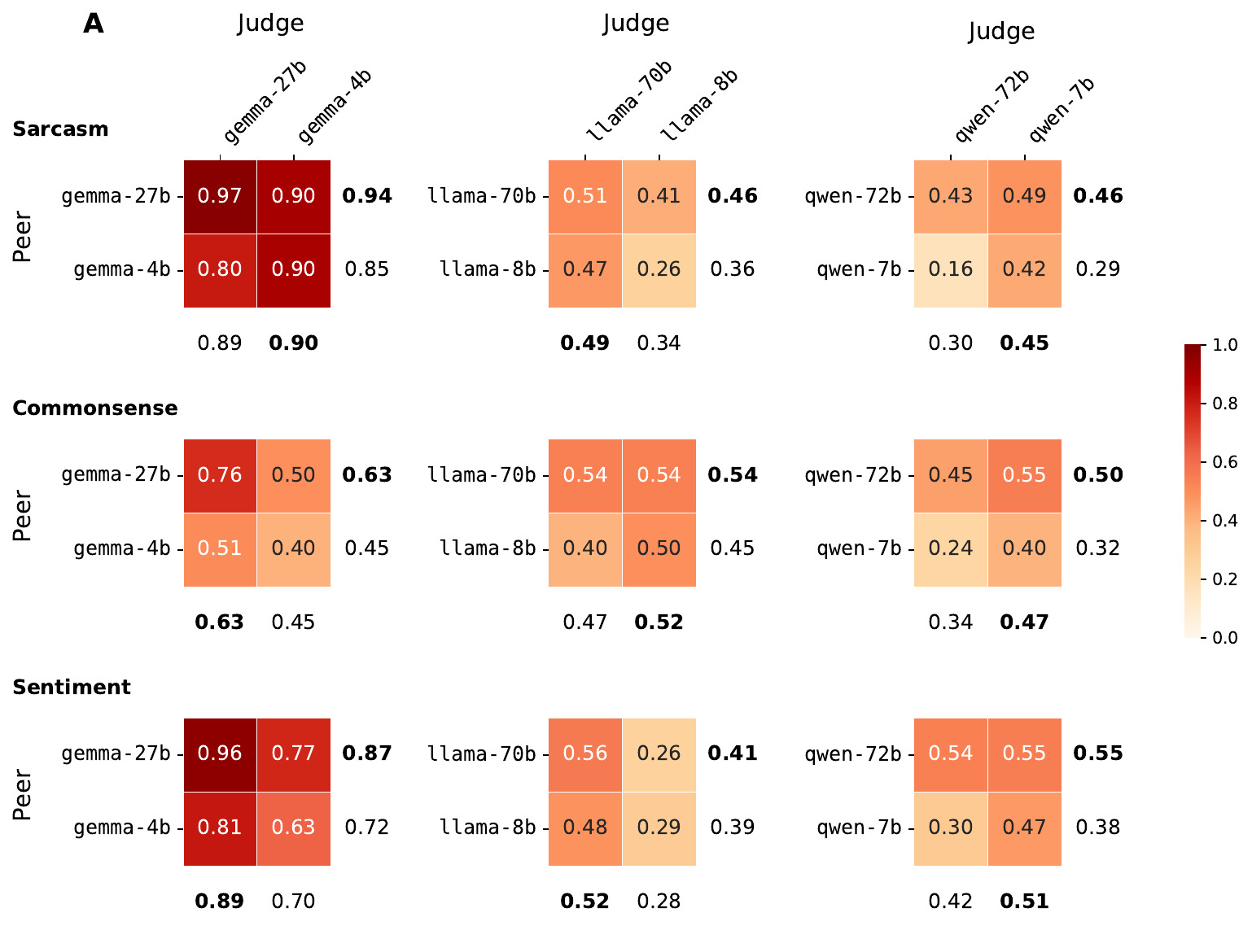}
    \includegraphics[width=0.49\textwidth]{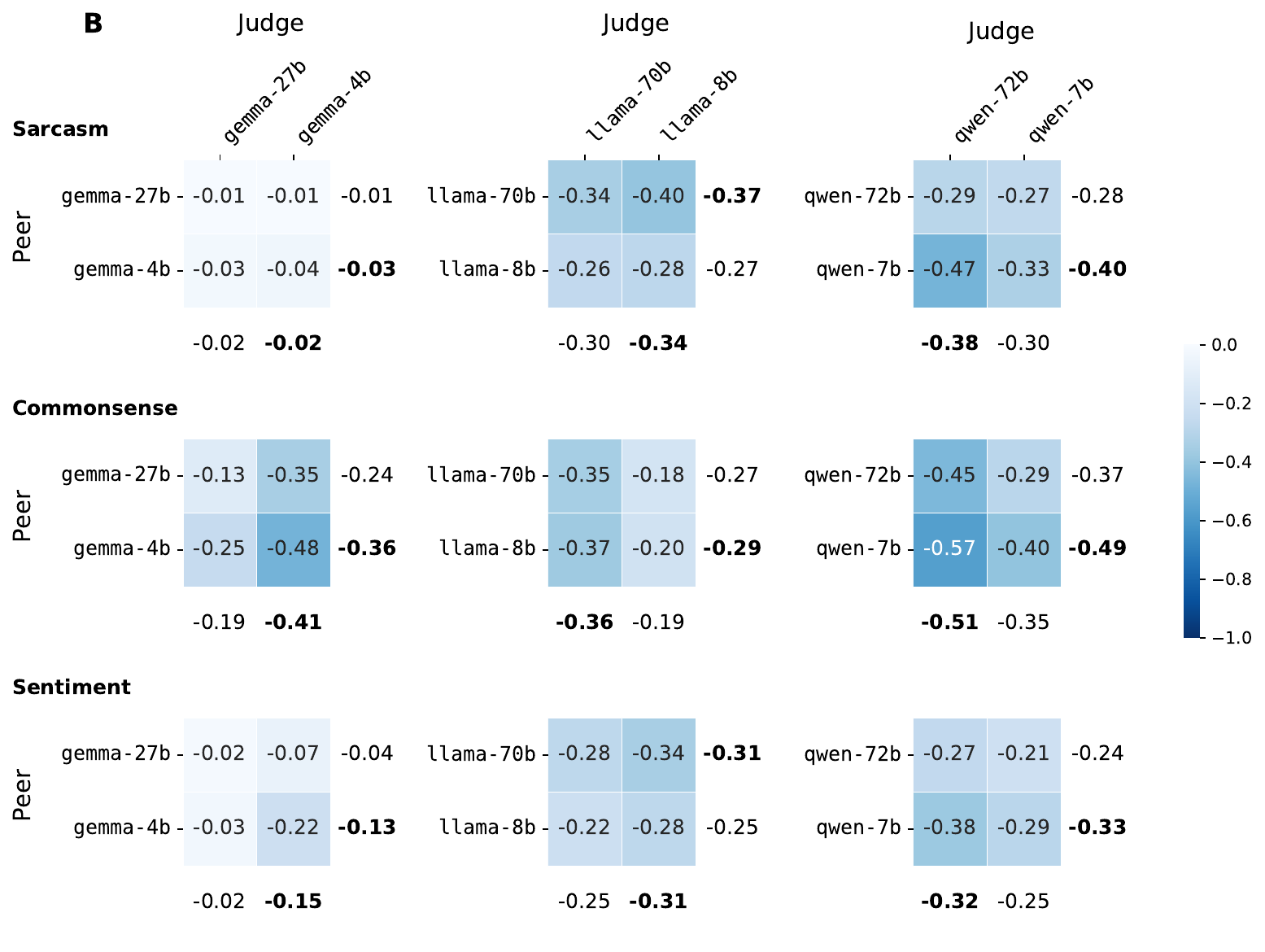}
    \caption{\textbf{Influence scores $\Delta_{\A|\B}$ across model pairs from the same family, with varying parameter size.} Positive influence scores are presented in panel A, backfiring scores in panel B. The anti-diagonal reports the outcomes of homogeneous systems. The largest absolute marginals in each row and column are in bold. The margin of error at $95\%$ confidence is $\leq$$0.01$ for all combinations of \judge, \peer, and dataset.}
    \label{fig:pairs-pos-overall}
\end{figure*}

To investigate the role of model size in persuasion outcomes, we focus on interactions between models of different size within the same family. On aggregate, larger models consistently achieve higher persuasiveness, as shown by the right marginal means in Figure~\ref{fig:pairs-pos-overall}A. This advantage, however, is driven mostly by their higher influence scores in homogeneous decision-making systems, and restricting the analysis to heterogeneous combinations makes the systematic size advantage disappear. In heterogeneous systems of \texttt{gemma} or \texttt{llama} agents, smaller models achieve comparable or higher influence. Smaller \texttt{qwen} models, by contrast, are not very effective \peers against large \texttt{qwen} \judges.

Backfiring follows a different pattern (Figure~\ref{fig:pairs-pos-overall}B): smaller \peers trigger stronger backfire overall. This is compatible with the intuition that the poorer justifications produced by smaller models are read by \judges as evidence for reinforcing their prior opinion. The effect is clearest in \texttt{gemma} and \texttt{qwen} systems, and weaker in \texttt{llama} systems, where the gap between sizes is less pronounced.

\begin{table*}[t]
\centering
\small
\begin{tabular}{l l c c >{\columncolor{white}}c c c >{\columncolor{white}}c}
\toprule
\multirow{2}{*}{\textsc{Experiment}} & \multirow{2}{*}{\textsc{Judge} $\boldsymbol{\leftarrow}$ \textsc{Peer}} & \multicolumn{3}{c}{$\Delta_{\A|\B}$} & \multicolumn{3}{c}{$\Delta^-_{\A|\B}$} \\
\cmidrule(lr){3-5} \cmidrule(lr){6-8}
& & \scriptsize \textsc{Main} & \scriptsize\textsc{Sensitivity} & \scriptsize Diff. & \scriptsize\textsc{Main} & \scriptsize \textsc{Sensitivity} & \scriptsize Diff. \\
\arrayrulecolor{black!80}\midrule
\textit{Temperature} & \texttt{llama-70b} $\leftarrow$ \texttt{qwen-72b-high} & 0.53 & 0.53 & 0.00 & -0.27 & -0.27 & 0.00 \\
& \texttt{llama-70b} $\leftarrow$ \texttt{qwen-72b-low} & 0.53 & 0.52 & -0.01 & -0.27 & -0.29 & -0.02 \\
& \texttt{qwen-72b} $\leftarrow$ \texttt{llama-70b-high} & 0.13 & 0.12 & -0.01 & -0.58 & -0.57 & +0.01 \\
& \texttt{qwen-72b} $\leftarrow$ \texttt{llama-70b-low} & 0.13 & 0.13 & 0.00 & -0.58 & -0.57 & +0.01 \\
\arrayrulecolor{gray!80}\midrule
\textit{Memoryless} & \texttt{llama-70b} $\leftarrow$ \texttt{qwen-72b} & 0.53 & 0.53 & 0.00 & -0.27 & -0.23 & +0.04 \\
\textit{Judge} & \texttt{qwen-72b} $\leftarrow$ \texttt{llama-70b} & 0.13 & 0.76 & +0.63 & -0.58 & -0.03 & +0.55 \\
\arrayrulecolor{gray!80}\midrule
\textit{No Explanation} & \texttt{llama-70b} $\leftarrow$ \texttt{qwen-72b} & 0.53 & 0.45 & -0.08 & -0.27 & -0.26 & +0.01 \\
& \texttt{qwen-72b} $\leftarrow$ \texttt{llama-70b} & 0.13 & 0.53 & +0.40 & -0.58 & -0.09 & +0.49 \\
\arrayrulecolor{gray!80}\midrule
\textit{Swap} & \texttt{llama-70b} $\leftarrow$ \texttt{gpt-20b} & 0.58 & 0.43 & -0.15 & -0.19 & -0.38 & -0.19 \\
& \texttt{gpt-20b} $\leftarrow$ \texttt{llama-70b} & 0.11 & 0.39 & +0.28 & -0.38 & -0.08 & +0.30 \\
\bottomrule
\end{tabular}
\caption{\textbf{Sensitivity analysis summary.} Influence scores $\Delta_{\A|\B}$ and backfiring scores $\Delta^-_{\A|\B}$, comparing results across all sensitivity analyses to the corresponding main experiment. Diff. = main - sensitivity.}
\label{tab:robustness-check}
\end{table*}

\subsection{Sensitivity analyses}

We perform four sensitivity analyses to examine the impact of model temperature and prompt design on the results. Because of computational constraints, we focus on the Sarcasm dataset and limit the experiments to the \texttt{qwen-72b} and \texttt{llama-70b} pair unless otherwise stated. We selected this pair for its comparable parameter size and sampling parameters, and for the asymmetry in the influence scores obtained in the main experiment, which makes it a particularly informative case for examining the stability of influence dynamics. All the prompts used in the sensitivity analysis are detailed in Section~\ref{app:sensitivity-prompts}. Table~\ref{tab:robustness-check} summarizes the resulting influence scores, which we detail next.

\vspace{4pt}\noindent\textit{Temperature.} We examine whether varying the \peerss sampling temperature affects \judge behavior. Prior work suggests that, in multiple-choice question answering, temperatures between $0.0$ and $1.0$ typically do not significantly affect accuracy, while performance can degrade beyond that range~\cite{temperature_renze}. We therefore vary the \peer temperature across $t \in \{0.2, \text{model default}, 1.2\}$, including $1.2$ to test whether a higher-variance setting, associated with more diverse and creative outputs, changes the \peerss ability to induce influence~\cite{temperature_creativity}. Influence scores show little to no difference across temperature configurations.

\vspace{4pt}\noindent\textit{Memoryless \judge.} We remove the \judgess own prior opinion from the prompt, so that in round 2 it is conditioned solely on the \peerss opinion. The two models respond quite differently. \texttt{Llama-70b} barely fluctuates, showing only a small decrease in backfiring relative to the main experiment. \texttt{Qwen-72b}, by contrast, shows a substantial increase in positive influence and a corresponding decrease in backfiring. Explicitly including the prior in the prompt therefore appears to mitigate susceptibility to peer influence.

\vspace{4pt}\noindent\textit{No explanations.} To isolate the contribution of the explanations, we remove them from both the \peer and the \judge outputs, keeping only the labels when prompting the \judge to reevaluate its decision in round 2. \texttt{Llama-70b} remains stable relative to the main experiment. \texttt{Qwen-72b} again behaves differently, with a higher positive influence score and reduced backfiring, mirroring the pattern observed in the memoryless condition. Access to the reasoning behind a label thus seems to contribute to model stability, and removing it leaves the model more vulnerable.

\vspace{4pt}\noindent\textit{Explanation swap.} We restructure the prompt so that the label-explanation pairs are swapped between the \peer and the \judge. This tests whether the \emph{content} of an explanation drives the final classification more than the \emph{identity} of the agent presenting it. We pair \texttt{llama-70b} with \texttt{gpt-20b}, as \texttt{gpt-20b} achieved by far the highest persuasiveness scores on Sarcasm, suggesting higher-quality explanations. The results indicate that content plays the major role. Exposed to its own reasoning as a \peer opinion, \texttt{llama-70b} becomes less susceptible, implying that its own explanations are less compelling as external input than \texttt{gpt-20b}'s. Symmetrically, \texttt{gpt-20b} becomes more susceptible and less prone to backfire when presented with its own reasoning. Both models therefore assign greater weight to \texttt{gpt-20b}'s label-explanation pairs regardless of the role from which they are presented, suggesting that \texttt{gpt-20b} produces more persuasive reasoning. A comparative analysis of the two models' explanations is provided in Section~\ref{app:qual}.

\section{Discussion} \label{sec:discussion}

Our findings confirm the strong effect of message passing on the decision-making outcomes of multi-agent systems. Influence patterns in heterogeneous systems depend on the specific combination of underlying language models and cannot be inferred from relative model size alone: small models can be effective persuaders and as impervious to influence as larger ones. The magnitude of the opinion shift is determined mostly by the model receiving the message and only to a lesser extent by the one exerting influence. It also depends on the properties of the target task, with larger shifts observed in harder tasks where decisions must rest on limited informative signal. Shifts can also occasionally reinforce the pre-existing agent opinion, similar to backfiring effects in human opinion dynamics~\cite{nyhan2010corrections, Cau2025SelectiveAgreementNotSychophancy}. Relative to their homogeneous counterparts, heterogeneous systems can either reduce or increase the variability of the final outcome, depending on model choice. In line with previous work emphasizing the gap between individual and collective agentic behavior~\cite{Ashery_2025, coppolillo2025unmasking}, we find that the classification certainty of an isolated agent does not imply stronger resistance to influence when the agent is moved to a multi-agent setting. This result runs against the widespread intuition that model certainty, estimated either from logits or from empirical generation probabilities, is a good proxy for confidence in the assigned label or for the correctness of that label~\cite{tian2023just, kang2026scalable}.

In this work, we deliberately focused on estimating influence rather than on classification performance against the ground-truth labels available for all datasets. For completeness, we report the difference in F1 scores between round 2 and round 1 in Section~\ref{app:f1_scores}. Three main findings emerge. First, peer interaction improves performance mostly on harder tasks, with Sarcasm detection benefiting the most, while performance on Sentiment generally declines in Round 2. Second, homogeneous \judge-\peer pairs perform worse than heterogeneous pairs. Third, systems with smaller models as \judges benefit more from interacting with larger models as \peers than the reverse.

Our findings come with three main experimental limitations. First, our framework considers only single-turn agent interactions on binary classification tasks. The single-turn setup isolates message-passing dynamics from the confounding factors that arise in larger and more articulated systems, and binary classification allows for an unambiguous quantitative definition of opinion shift. These constraints nevertheless call for broader exploration of emergent decision-making patterns in large, complex systems of self-organized AI agents~\cite{flint2026group} working on open-ended tasks. Second, our results rest on a limited pool of open-source, quantized models tested over a limited range of sampling parameters. Despite efforts to maximize variety, the findings are not guaranteed to generalize to other LLMs or to very large commercial models. Quantization might also affect the outcomes in way that might be hard to predict~\cite{ferrara2026quantibias}. Last, our analysis weights all configurations of influence equally in the final estimate, a deliberate choice to examine every configuration in equal detail. However, the empirical distribution is concentrated in a subset of configurations, as shown in Section~\ref{app:match-type-scores}.

Despite these limitations, our results carry concrete implications for the design of multi-agent decision-making systems. The choice of which combination of models to use emerges as a first-order design decision requiring systematic evaluation. The model mix determines not only the system's inference capabilities~\cite{ye2025xmasbuildingmultiagentsystems} but also its stability, affecting its propensity to overturn the decisions that its agents would reach in isolation. Instability also increases with intrinsic task uncertainty, making system evaluation necessarily task-specific: model combinations that are stable on easy tasks provide no guarantee of stability on harder ones. Such instability can also become a security liability. In open application scenarios, where multiple parties contribute resources to a shared workflow, the system's outcome can be reversed effectively through the injection of a contrarian agent powered by a relatively small model~\cite{flint2026indirect}. In other words, adversarial capability is not gated by model scale, and the properties that make a system robust to attacks cannot be estimated from the performance of its components in isolation. Design practices that select agents on individual capability alone (such as their individual confidence or performance on the task) therefore leave a substantial attack surface exposed, and should be revisited in light of these findings to build more reliable and robust multi-agent systems.

\section{Methods}

\subsection{Influence estimation}

\begin{figure}[t!]  
    \includegraphics[width=\columnwidth]{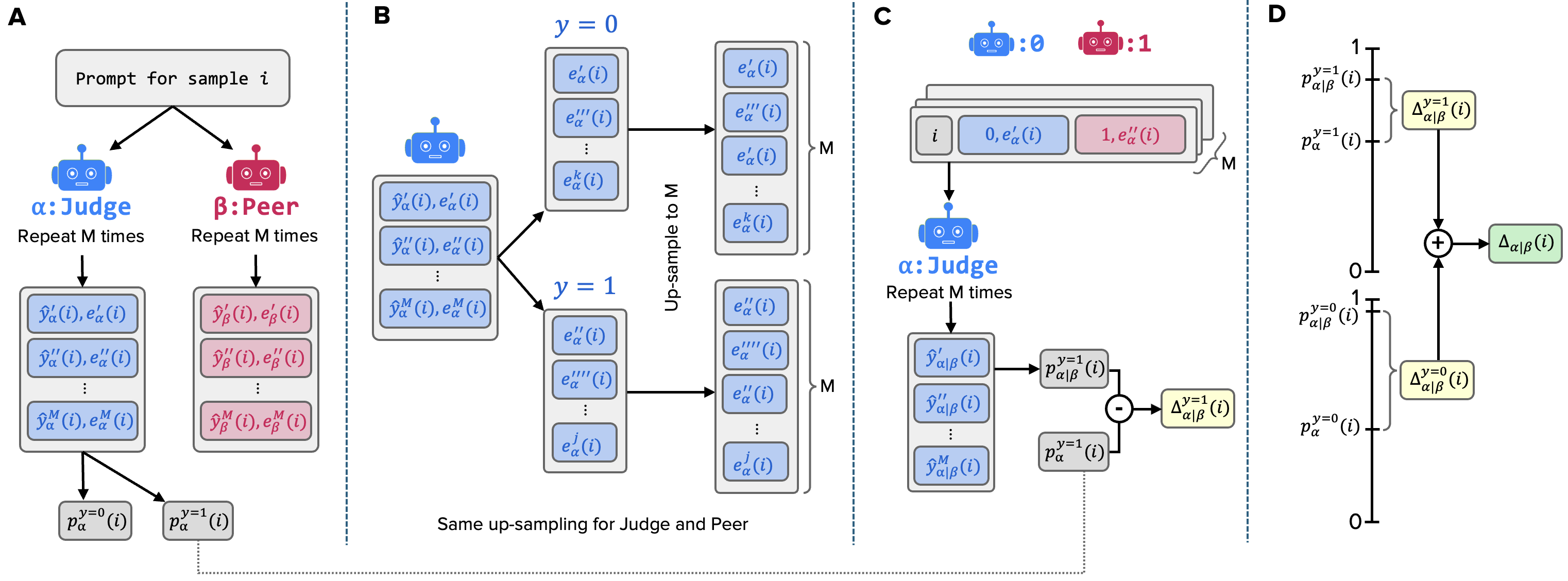}
    \caption{\textbf{Overview of the framework}. (A) In round 1, the \judge and \peer independently generate $M$ outputs for the same sample $i$. The \judgess probabilities of generating each output label is calculated. (B) Outputs from round 1 are upsampled so that both \judge and \peer have $M$ explanations for each distinct label they generated in round 1. (C) $M$ instances of contrasting labels-explanations pairs from the \judge and \peer are given in input to the \judge in round 2, resulting in $M$ new labels, out of which a new label probability is calculated. The comparison between the label probability in round 1 and that in round 2 yields the $\Delta$ value. (D) The sum of $\Delta$ values for both directions of influence ($0 \rightarrow 1$ and $1 \rightarrow 0$) is the final influence score.}
    \label{fig:diagram}
\end{figure} 

Figure~\ref{fig:diagram} illustrates the different steps of the framework. In the first round, the \judge $\A$ and the \peer $\B$ are independently prompted to generate a binary classification label $\hat{y}$ for the same text sample $i$, along with a natural language explanation $e$ motivating their decision:
\begin{equation}
\A(i) \rightarrow \langle \hat{y}_{\A}(i), e_{\A}(i) \rangle; \quad \B(i) \rightarrow \langle \hat{y}_{\B}(i), e_{\B}(i) \rangle.
\label{eq:round1}
\end{equation}
In the second round, the \peerss output is passed to the \judge, which is asked to reconsider the label for sample $i$ given: i) its own round-1 output, and ii) the label and explanation from the \peer.
\begin{equation}
\A(i, \hat{y}_{\A}(i), e_{\A}(i), \hat{y}_{\B}(i), e_{\B}(i)) \rightarrow \hat{y}_{\A|\B}(i).
\label{eq:round2}
\end{equation}
The label $\hat{y}_{\A|\B}(i)$ is the final verdict of the \judge $\A$ after being conditioned by the \peer $\B$. Because the second round is the last one and no further interaction between agents is required, the final label is produced without an explanation.
To account for the stochasticity of model outputs, we assess the influence of $\B$ over $\A$ by repeating both rounds $M$ times and comparing the shift in the empirical Bernoulli distributions over the binary label across rounds.
Specifically, we estimate the probability of $\A$ independently choosing label $y$ over $M$ trials of round 1:
\begin{equation}
    p_\A^y(i) = \frac{1}{M} \sum_{m=1}^M \mathbb I(\hat{y}_\A^m(i)=y),
    \label{eq:pik}    
\end{equation}
By definition, $p_\A^y(i) = 1 - p_\A^{\neg y}(i)$. We than calculate the probability of producing label $y$ over $M$ trials of round 2, when the information from the \peer agent $\B$ is taken into account:
\begin{equation}
    p_{\A|\B}^y(i) = \frac{1}{M} \sum_{m=1}^M \mathbb I(\hat{y}^m_{\A|\B}(i)=y).
    \label{eq:cond_pik}
\end{equation}
Here, $\hat{y}^m_\bullet$ denotes the label produced in the $m^{th}$ trial of the round, and $\mathbb{I}(\cdot)$ is the indicator function. We note that, unlike in round 1, $p_{\A|\B}^y(i) \neq 1 -p_{\A|\B}^{\neg y}(i)$, because the two probabilities for $y$ and $\neg y$ are obtained from two independent experiments with M trials each: the first where the \judge starts from $\neg y$ and it is influenced toward $y$, and the second from $y$ toward $\neg y$. The difference between the values in Equations~\ref{eq:pik} and~\ref{eq:cond_pik} expresses the \textit{shift} in the probability of $\A$ choosing label $y$ due to the influence of $\B$:
\begin{equation}
    \Delta_{\A|\B}^{y}(i) = p_{\A|\B}^{y}(i) - p_\A^{y}(i).
\end{equation}
The shift is positive when $\B$ causes an increase in the probability of $\A$ generating label $y$, and negative when the message from $\B$ leads $\A$ to reinforce its preference on the label selected at round 1. To avoid conflating positive and negative shifts into a single score, we tally them separately and call the positive scores \emph{influence scores} and the negative scores \emph{backfiring scores}. Alongside these scores, we report the probability that backfiring occurs.
The total influence of $\B$ over $\A$ on instance $i$ is the sum of the positive shifts toward the two possible labels (from $\neg y$ to $y$ and from $y$ to $\neg y$):
\begin{equation}
    \Delta_{\A|\B}(i) = \Delta_{\A|\B}^{y}(i) + \Delta_{\A|\B}^{\neg y}(i)
    \label{eq:influence_sample}
\end{equation}
Values of $\Delta_{\A|\B}(i)$ range from 0 to 1. A value of zero indicates that $\B$ has no influence over $\A$, as the label distribution is unchanged after the interaction; a value of one indicates that $\A$ always revises its initial decision after the interaction.
Last, the \emph{influence score} of $\B$ over $\A$ is calculated as an average over the outcomes of all $N$ instances:
\begin{equation}
    \Delta_{\A|\B} = \frac{1}{N} \sum_{i=1}^N \Delta_{\A|\B}(i).
    \label{eq:influence}
\end{equation}
To calculate the standard error for $\Delta_{\A|\B}$, we start by calculating the standard error of the probabilities $p_{\A}^y(i)$ and $p_{\A|\B}^y(i)$ using the formula for the Bernoulli distribution:
\begin{equation}
    SE(p) = \sqrt{\frac{p(1-p)}{M}}.
    \label{eq:standard_error}
\end{equation}
We then propagate the error up to the calculation of $\Delta_{\A|\B}$.

\subsection{Match Cases}

In round 2, the label-explanation pairs of the \judge and the \peer are matched so that their labels oppose each other, covering both influence directions $y \rightarrow \neg y$ and $\neg y \rightarrow y$. Whether such matches can be formed depends on the probabilities with which $\A$ and $\B$ generate each label for the target sample $i$ in round 1. For example, if $\A$ always outputs $y=0$ ($p^{y=0}_\A(i)=1$) and $\B$ always outputs $y=1$ ($p^{y=1}_\B(i)=1$), influence can be tested only in the direction $0 \rightarrow 1$, a \emph{match case} we denote as $0$:$1$. If instead both agents have non-zero probability of generating both labels in round 1, both directions can be tested, and we denote this case as $B$:$B$, where $B$ marks an agent that produces both labels. The complete set of cases that enable disagreeing interactions in round 2 is $B$:$1$, $B$:$0$, $0$:$B$, $1$:$B$, $1$:$0$, $0$:$1$, and $B$:$B$, and every instance $i$ falls into exactly one of them. For $1$:$0$ and $0$:$1$ we test a single direction of influence on all $M$ available instances. In all other cases, we upsample the label-explanation pairs to $M$ for each available direction, so that every case yields $M$ repetitions in round 2 (Figure~\ref{fig:diagram}B illustrates this construction). Upsampling, rather than simply increasing the value of $M$, is necessary because the label distribution is inherently skewed for some model-sample pairs. It also keeps the number of repetitions consistent across cases, ensuring that the influence measure rests on the same number of interactions. In a homogeneous setting, where both \judge and \peer are instantiated from the same underlying language model, disagreeing output pairs can be constructed only in the $B$:$B$ case, in which the model produced mixed labels in its round-1 predictions.
Finally, to weight all valid label combinations equally when estimating the influence between a model pair, we compute the influence score of Equation~\ref{eq:influence} separately for each match case and then macro-average across cases. Only for the match cases $B$:$1$ and $B$:$0$, $\Delta_{\A|\B}(i)$, as defined in Equation~\ref{eq:influence_sample}, is strictly lower than 1 by definition, since the \judgess certainty in isolation is $<$$1$ and the \peer can steer the decision in only one direction (toward 1 for $B$:$1$, toward 0 for $B$:$0$). We therefore min-max normalize $\Delta_{\A|\B}(i)$ to bring it into the desired $[0,1]$ range. To keep computational cost manageable, we cap each match case at $7{,}000$ samples. Given $M=10$, this amounts to $70{,}000$ generations in total for each combination of model, match case, and dataset. The exact cardinality of all match cases is reported in Section~\ref{app:match-type-scores}.

As a robustness check, we also test agreeing interactions where both the \judge and the \peer output the same label in round 1, and find that almost no opinion shifts occurs, as expected (Section~\ref{app:agreeing-influence}).

\section*{Code and Data Availability}

Code and data are publicly available at \url{https://github.com/fridanl/InfluenceDynamicsMAS}.

\section*{Acknowledgments}

L.M.A. acknowledges the support from the Horizon-CL4 project MINGLE (EU grant: 101341932), and from the Carlsberg Foundation through the COCOONS project (CF21-0432). R.P.-S. acknowledges financial support from project PID2022-137505NB-C21/AEI/10.13039/501100011033/FEDER UE and the Acadèmia d’Excel·lència Award, funded by the Generalitat de Catalunya. We thank Stian Sivertsgård for his exploratory work during the conceptualization phase of the project.

\bibliographystyle{ACM-Reference-Format}
\bibliography{bibliography}

\clearpage

\appendix
\input{appendix}

\end{document}

%% file: appendix.tex
\setcounter{section}{0}
\setcounter{figure}{0}
\setcounter{table}{0}
\renewcommand{\thesection}{S\arabic{section}}
\renewcommand{\thetable}{S\arabic{table}}
\renewcommand{\thefigure}{S\arabic{figure}}

\section*{SUPPLEMENTARY INFORMATION}

\section{Data preprocessing}\label{app:preprocessing}

We preprocess the Sarcasm and Sentiment datasets to remove low-quality samples. The Commonsense dataset is already well curated and required no additional steps. The preprocessing scripts are published in the project repository (\texttt{src/preprocessing.py}).

\vspace{4pt} \noindent \textit{Sarcasm.} We omit all claims shorter than $6$ words to ensure sufficient semantic context. We also apply a keyword-based filter to remove claims touching on sensitive topics such as \emph{firearms}, \emph{holocaust}, and \emph{suicide}, so that model outputs are not affected by guardrails. The full keyword list is provided in the project repository.

\vspace{4pt} \noindent \textit{Sentiment.} Because of its tree-based structure, the Sentiment dataset contains many semi-duplicate instances sharing most of their text, with only a few additional words appended. We remove semi-duplicates that carry the same label and retain those in which the added text changes the sentiment. Consider the following three labeled samples:
\begin{enumerate}[leftmargin=*]
    \item \texttt{\textbf{[positive]} The film boasts at least a few good ideas and features some decent performances.}
    \item \texttt{\textbf{[positive]} The film boasts at least a few good ideas and features some decent performances, but,}
    \item \texttt{\textbf{[negative]} The film boasts at least a few good ideas and features some decent performances, but the result is disappointing.}
\end{enumerate}
The first two express the same sentiment and are treated as semi-duplicates with identical labels, so we keep only the longer text. The third is retained, since the added phrase flips the sentiment from positive to negative.

\section{Model Specifications}\label{app:model-specifications}

In \autoref{tab:model-specs}, we provide further details about the models used in our experiments. Further details can be obtained on their original Hugging Face repository pages.

\begin{table*}[h]
    \centering
    \small
    \setlength{\tabcolsep}{8pt}
    \begin{tabular}{l l l c }
        \toprule
        \textsc{Model name} & \textsc{Provider} & \textsc{Repository} ID & \textsc{Sampling params.} \\
        \arrayrulecolor{black!80}\midrule
        \texttt{gemma-3-4b}     &   Google          &   unsloth/gemma-3-4b-it-unsloth-bnb-4bit      &   $\times$ \\
        \texttt{gemma-3-27b}    &   Google          &   unsloth/gemma-3-27b-it-bnb-4bit             &   $\times$ \\
        \texttt{llama-3.1-8b}   &   Meta            &   unsloth/Meta-Llama-3.1-8B-Instruct-bnb-4bit &   \checkmark \\
        \texttt{llama-3.3-70b}  &   Meta            &   unsloth/Llama-3.3-70B-Instruct-bnb-4bit     &   \checkmark \\ 
        \texttt{qwen-2.5-7b}    &   Alibaba Cloud   &   unsloth/Qwen2.5-7B-Instruct-bnb-4bit        &   \checkmark \\
        \texttt{qwen-2.5-72b}   &   Alibaba Cloud   &   unsloth/Qwen2.5-72B-Instruct-bnb-4bit       &   \checkmark \\
        \texttt{gpt-oss-20b}    &   OpenAI          &   unsloth/gpt-oss-20b-unsloth-bnb-4bit        &   
        $\times$ \\
        \bottomrule
    \end{tabular}  
    \caption{\textbf{Specification of models}. ``Sampling params'' refer to whether a generation configuration-file specifying default sampling parameters is provided in the repository.}
    \label{tab:model-specs}
\end{table*}

\vspace{4pt} \noindent \textit{Sampling parameters.} None of the two \texttt{gemma} models have any sampling parameters specified in their Hugging Face repositories, neither in the unsloth version or the repository for the original model. We therefore followed the recommendations from online sources  discussing sampling parameters: \url{https://www.reddit.com/r/LocalLLaMA/comments/1j9hsfc/gemma_3_ggufs_recommended_settings/} and \url{https://docs.aimlapi.com/api-references/text-models-llm/google/gemma-3} (last accessed 30/09/2026). \texttt{Gpt-20b} does not specify sampling parameters in its Hugging Face repository. We therefore use the parameters from OpenAI's official \texttt{gpt-oss} Github repository, stated in the model's README (\url{https://github.com/openai/gpt-oss/tree/main}, last accessed 30/09/2026).

\section{Prompts}\label{app:prompts}

\begin{figure*}[h!]  
    \centering
    \includegraphics[width=0.5\textwidth]{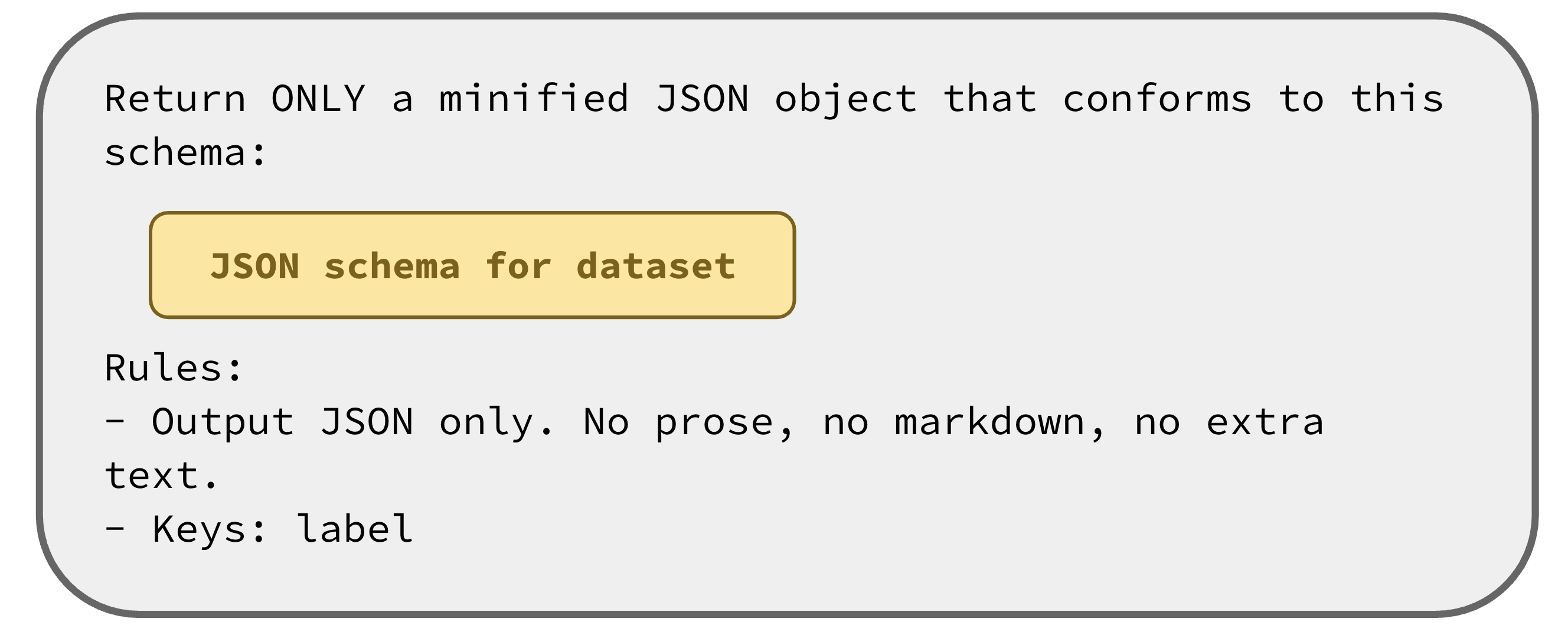}
    \caption{\textbf{System prompt for all model calls.}}
    \label{fig:prompt-system}
\end{figure*}

\begin{figure*}[h!]  
    \centering
    \includegraphics[width=1.0\textwidth]{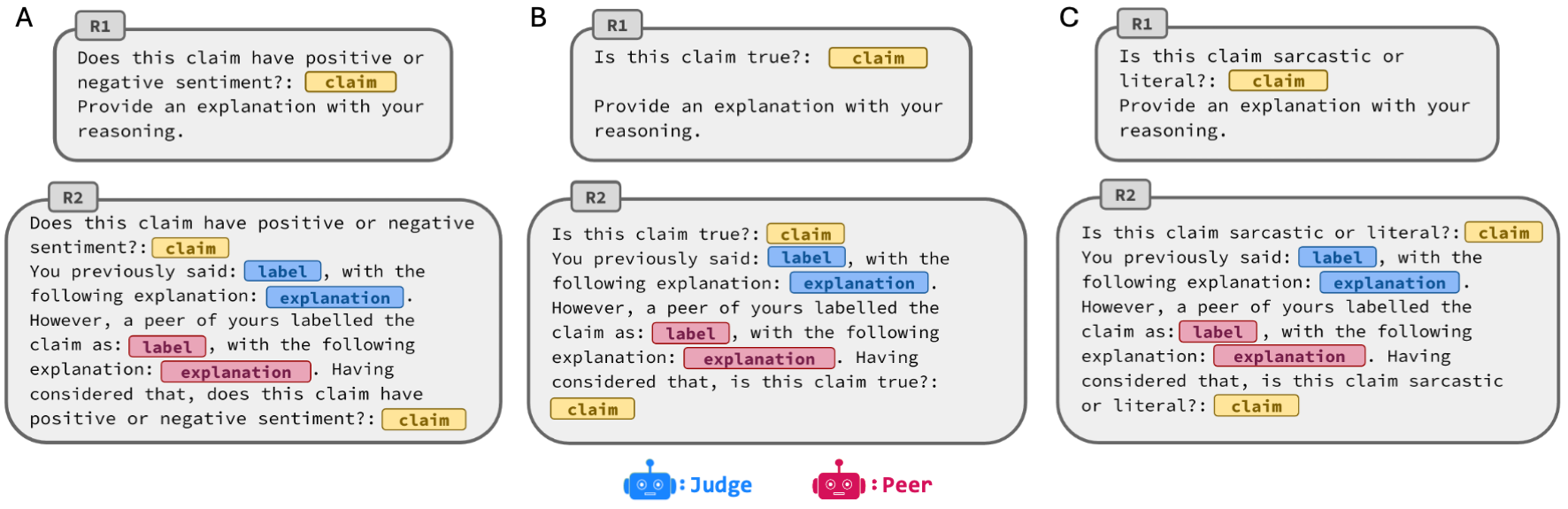}
    \caption{\textbf{Round 1 and round 2 prompts.} Slightly different prompts are used for Sentiment (A) Commonsense (B), and Sarcasm (C). In Round 2, colors represent the label and explanation pairs coming from the \judgess (blue) and \peerss (red) outputs produced in round 1.}
    \label{fig:prompt-tasks}
\end{figure*}

System prompt and user prompt structures for all datasets for both rounds are presented in Figures~\ref{fig:prompt-system} and~\ref{fig:prompt-tasks}.

\section{Discarded claims from invalid outputs}\label{app:discarded-claims}

Table~\ref{tab:discarded-claims} reports the number of round-1 instances discarded because their formatting did not conform to the expected JSON structure. Given  their very small number, we exclude them from all further processing.

\clearpage

\begin{table}[t!]
\small
\setlength{\tabcolsep}{3pt}
\centering
\begin{tabular}{lccc}
\toprule
\shortstack{\textsc{Model} \\ $\quad$ }
& \shortstack{\textsc{Sarcasm}  \\ \tiny $N=137,454$} 
& \shortstack{\textsc{Commonsense}  \\ \tiny $N=9,264$} 
& \shortstack{\textsc{Sentiment} \\ \tiny $N=91,964$ }\\
\midrule
\texttt{gemma-27b} &  4 & 1 & 11 \\
\texttt{gemma-4b}  & 0 & 8 & 43 \\
\texttt{llama-70b} & 1 & 12 & 59 \\
\texttt{llama-8b}  & 7 & 44 & 5 \\
\texttt{qwen-72b}  & 27 & 1 & 0\\
\texttt{qwen-7b}   & 21 & 31 & 18 \\
\texttt{gpt-20b}   & 0 & 0 & 0 \\
\bottomrule
\end{tabular}
\caption{\textbf{Data cleaning.} Number of instances discarded from baseline outputs per model and dataset. $N$ denotes the total dataset size.}
\label{tab:discarded-claims}
\end{table}

\section{Influence heatmaps}\label{app:influence-heatmaps}

\begin{figure*}[h!]  
    \centering
    \includegraphics[width=1\textwidth]{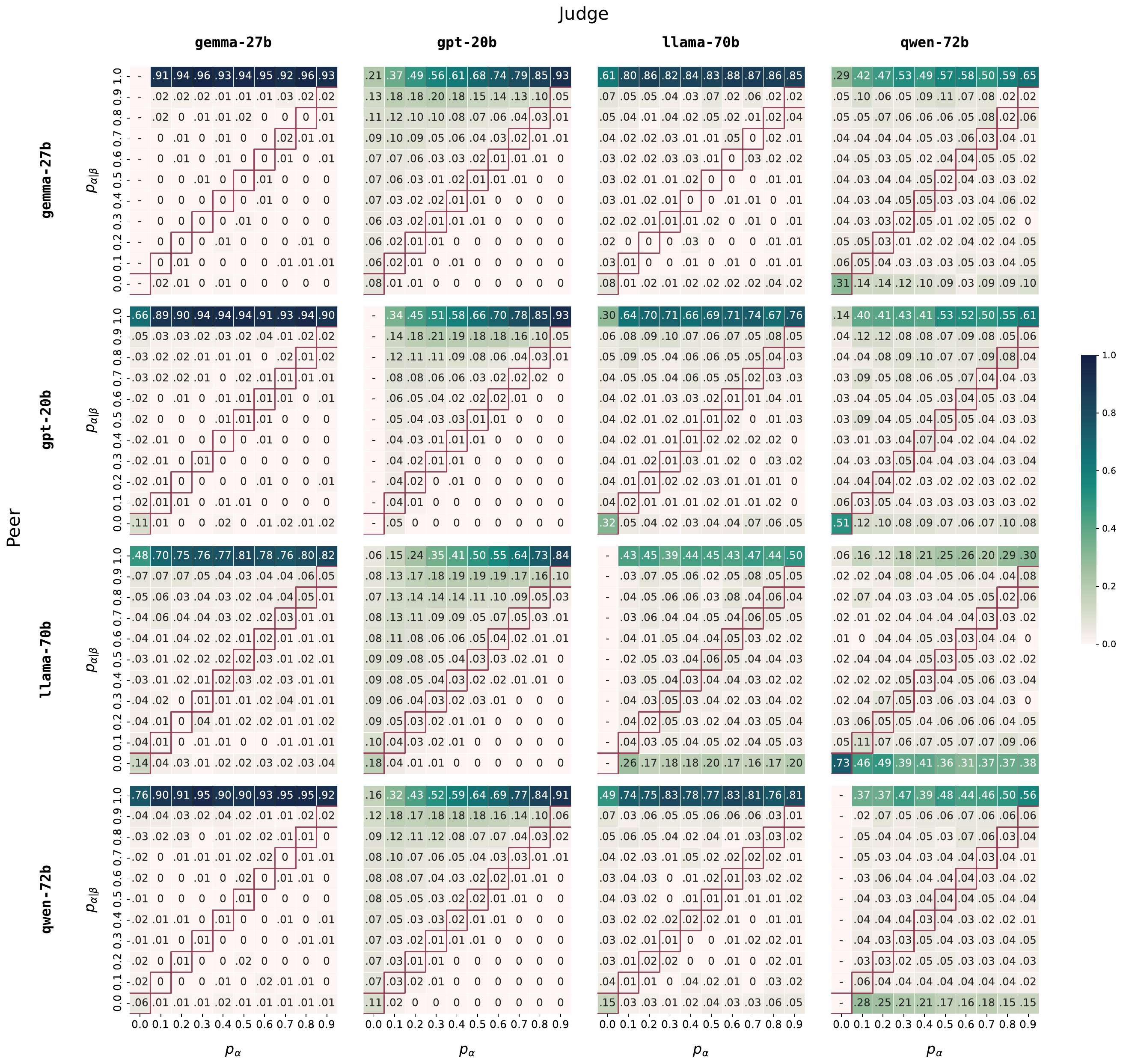}
    \caption{\textbf{Conditional distribution of post-interaction certainty ($p_{\A|\B}$) given individual model certainty $p_{\A}$ for each receiver-peer pair on the Sentiment dataset}. Plots on the diagonal correspond to homogeneous systems. The leftmost column in diagonal plots is empty since samples with $p_{\A}=0$ in homogeneous systems cannot generate any disagreeing pairs for round 2.}
    \label{fig:influence-heatmaps-sentiment}
\end{figure*}

\begin{figure*}[t!]  
    \centering
    \includegraphics[width=1\textwidth]{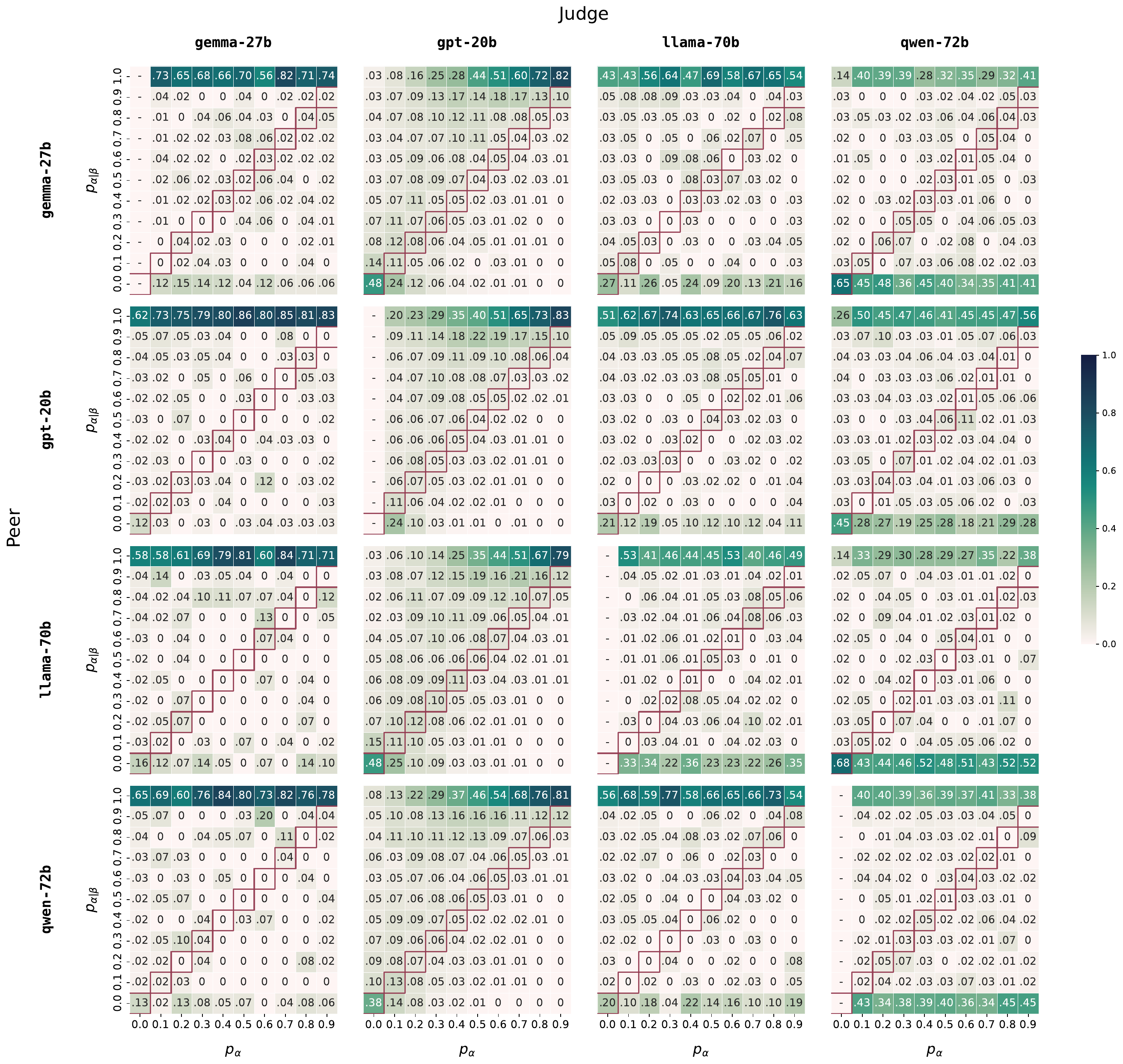}
    \caption{\textbf{Conditional distribution of post-interaction certainty ($p_{\A|\B}$) given individual model certainty $p_{\A}$ for each receiver-peer pair on the Commonsense dataset}. Plots on the diagonal correspond to homogeneous systems. The leftmost column in diagonal plots is empty since samples with $p_{\A}=0$ in homogeneous systems cannot generate any disagreeing pairs for round 2.}
    \label{fig:influence-heatmaps-commonsense}
\end{figure*}

\begin{figure*}[t!]  
    \centering
    \includegraphics[width=1\textwidth]{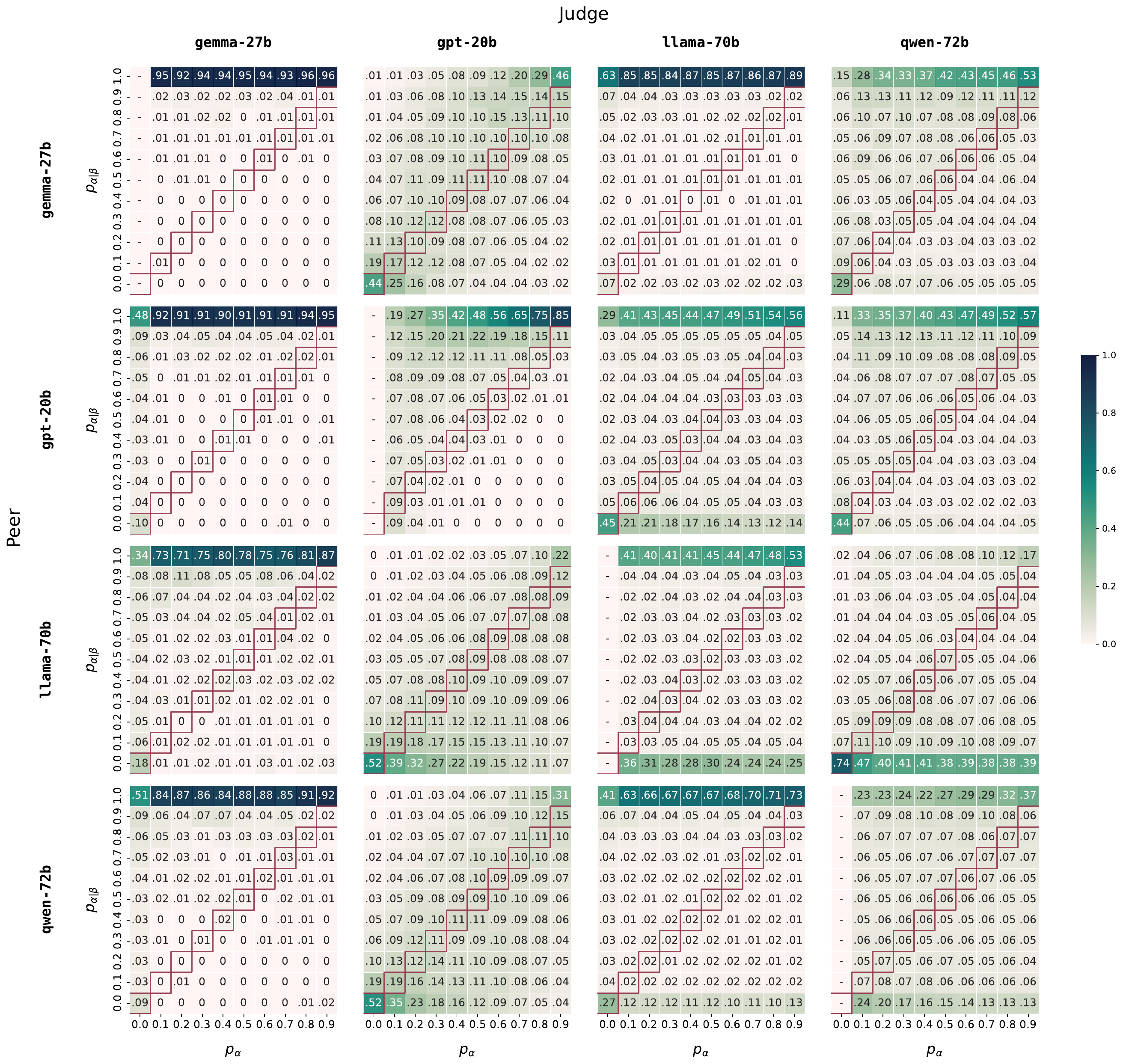}
    \caption{\textbf{Conditional distribution of post-interaction certainty ($p_{\A|\B}$) given individual model certainty $p_{\A}$ for each receiver-peer pair on the Sarcasm dataset}. Plots on the diagonal correspond to homogeneous systems. The leftmost column in diagonal plots is empty since samples with $p_{\A}=0$ in homogeneous systems cannot generate any disagreeing pairs for round 2.}
    \label{fig:influence-heatmaps-sarcasm}
\end{figure*}

By inspecting the distribution of post-interaction certainty ($p_{\A|\B}$) conditional on individual model certainty ($p_{\A}$), we assess whether a model's classification certainty in isolation correlates with its likelihood of changing label after being influenced. The conditional distributions for the three datasets are presented in Figures~\ref{fig:influence-heatmaps-sentiment},~\ref{fig:influence-heatmaps-commonsense}, and~\ref{fig:influence-heatmaps-sarcasm}. Probability mass concentrated on the diagonal (marked in red) indicates that the \judgess post-interaction certainty mirrors its baseline, signaling stability under influence. Mass above the diagonal reflects positive influence, and mass below it represents backfiring. \texttt{Gpt-20b} is the only model showing appreciable correlation between $p_{\A}$ and $p_{\A|\B}$; in all others, the conditional mass concentrates at $p_{\A|\B}=0$ or $p_{\A|\B}=1$. In all model combinations, the probability mass on the diagonal is low, indicating a general instability of all models.

\clearpage

\clearpage

\clearpage

\section{Influence scores on agreeing interactions} \label{app:agreeing-influence}

\begin{figure*}[h]  
    \centering
    \includegraphics[width=0.6\textwidth]{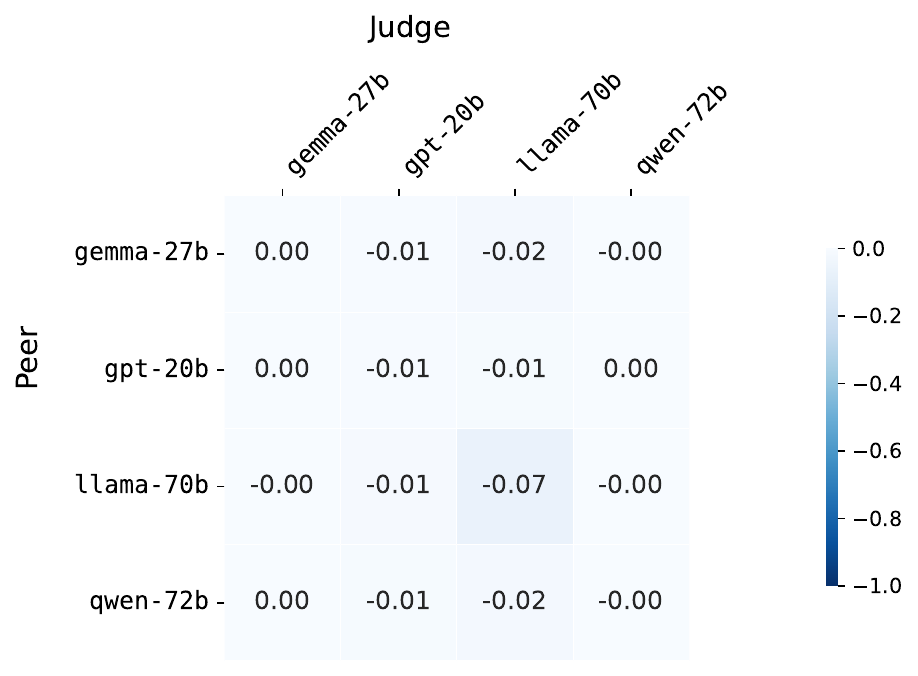}
    \caption{\textbf{Backfiring scores for agreeing interactions across model pairs on the Sarcasm dataset}. Columns shows how susceptible a receiver is to influence from the different peers, while the rows show how much influence a peer can exert on different receivers.}
    \label{fig:large-neg-overall-agreeing}
\end{figure*}

In Figure~\ref{fig:large-neg-overall-agreeing}, we report results for agreeing interactions on the Sarcasm dataset only, as a robustness check. All scores are based on instances in the $\text{0:0}$ and $\text{1:1}$ match cases. In these cases, the \judge already assigns all probability mass to the \peerss label, leaving no room for positive influence, so any shift can register only as backfiring. When the \peer proposes a label matching the \judgess opinion, models remain stable and interactions yield near-zero backfiring.

\clearpage

\section{Backfiring probability} \label{app:backfiring-probability}

\begin{figure*}[h]  
    \centering
    \includegraphics[width=\textwidth]{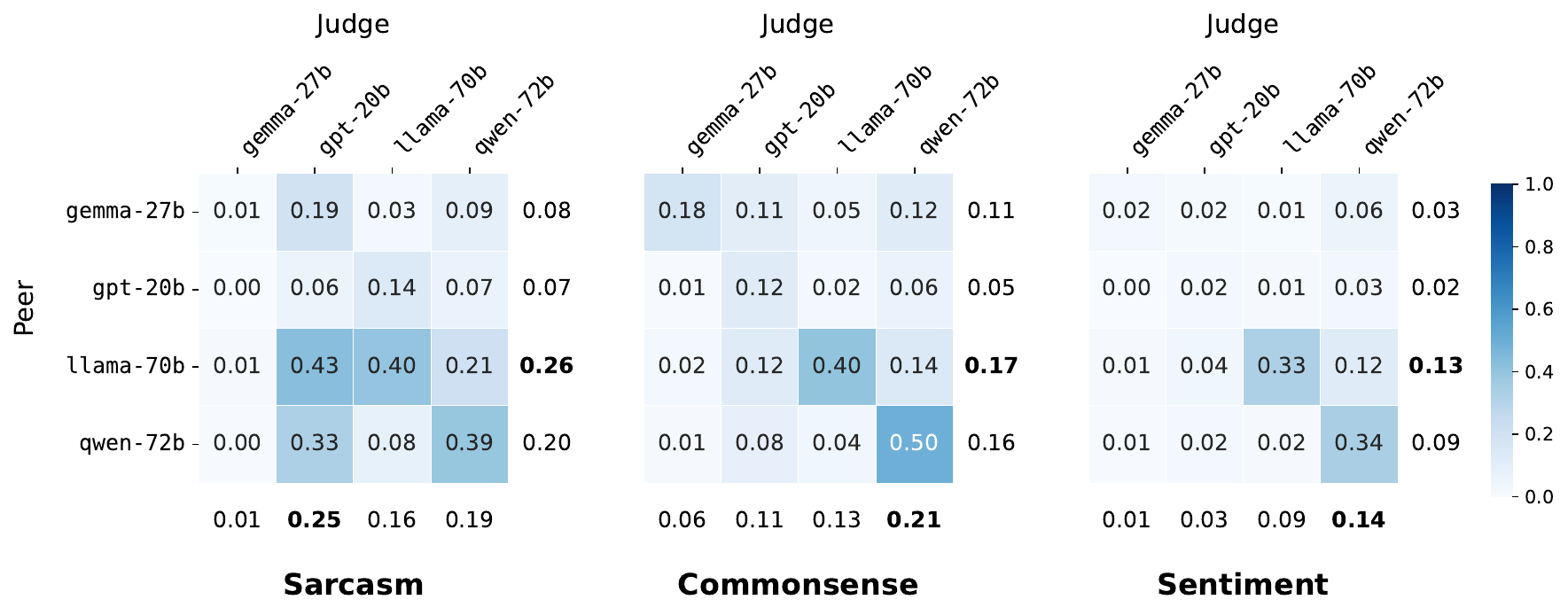}
    \caption{\textbf{Probability of backfiring across model pairs and datasets.}}
    \label{fig:prob_backfiring_inter}
\end{figure*} 

\begin{figure*}[h]  
    \centering
    \includegraphics[width=0.8\textwidth]{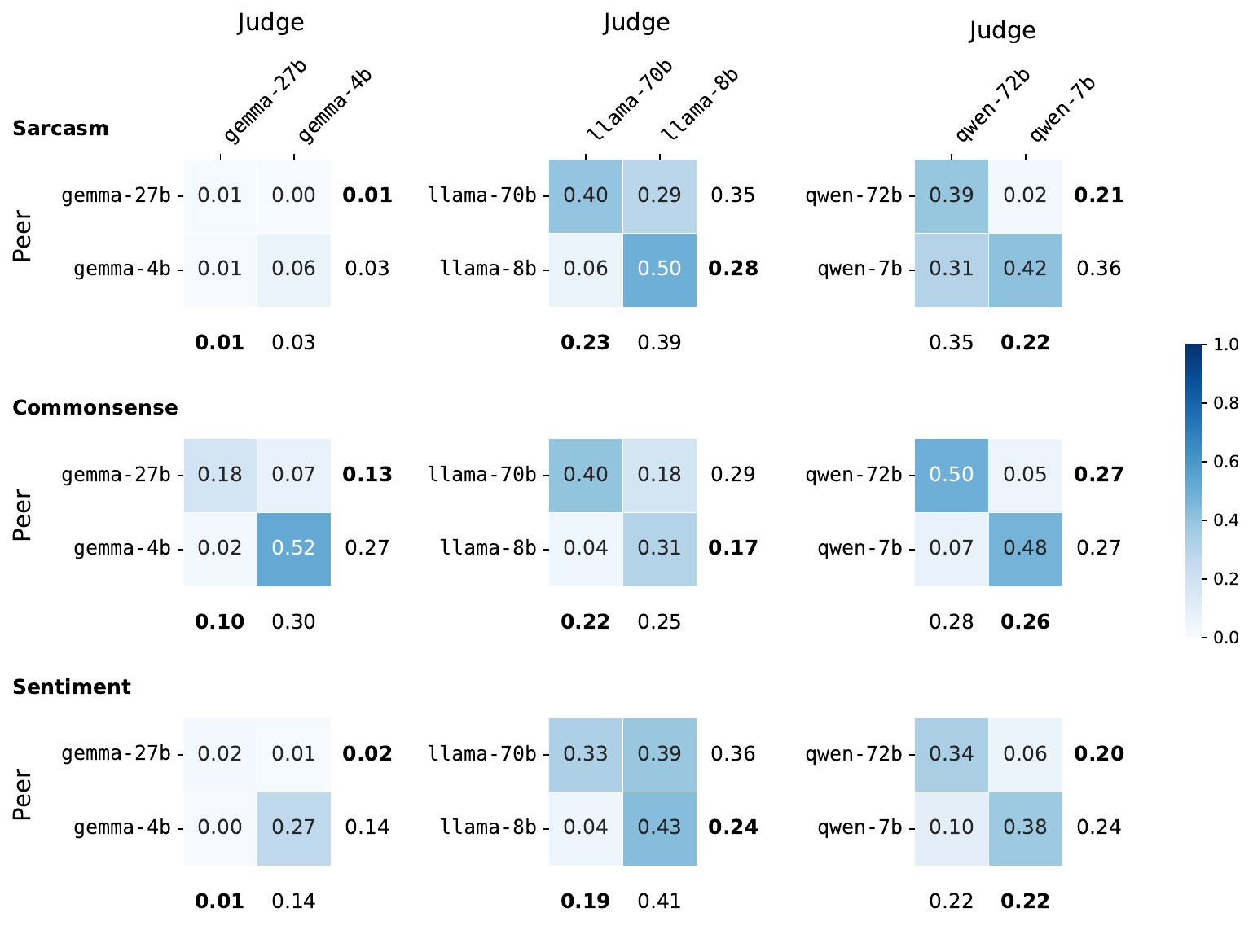}
    \caption{\textbf{Probability of backfiring across model pairs from the same family, with varying parameter size.}}
    \label{fig:prob_backfiring_intra}
\end{figure*} 

Figures~\ref{fig:prob_backfiring_inter} and~\ref{fig:prob_backfiring_intra} report the probability that a peer induces backfiring in judges.

\clearpage

\section{Influence Comparison between Homogeneous and Heterogeneous Systems} \label{app:homo-vs-hetero}

\begin{figure*}[htbp]
    \centering
    \includegraphics[width=1\textwidth]{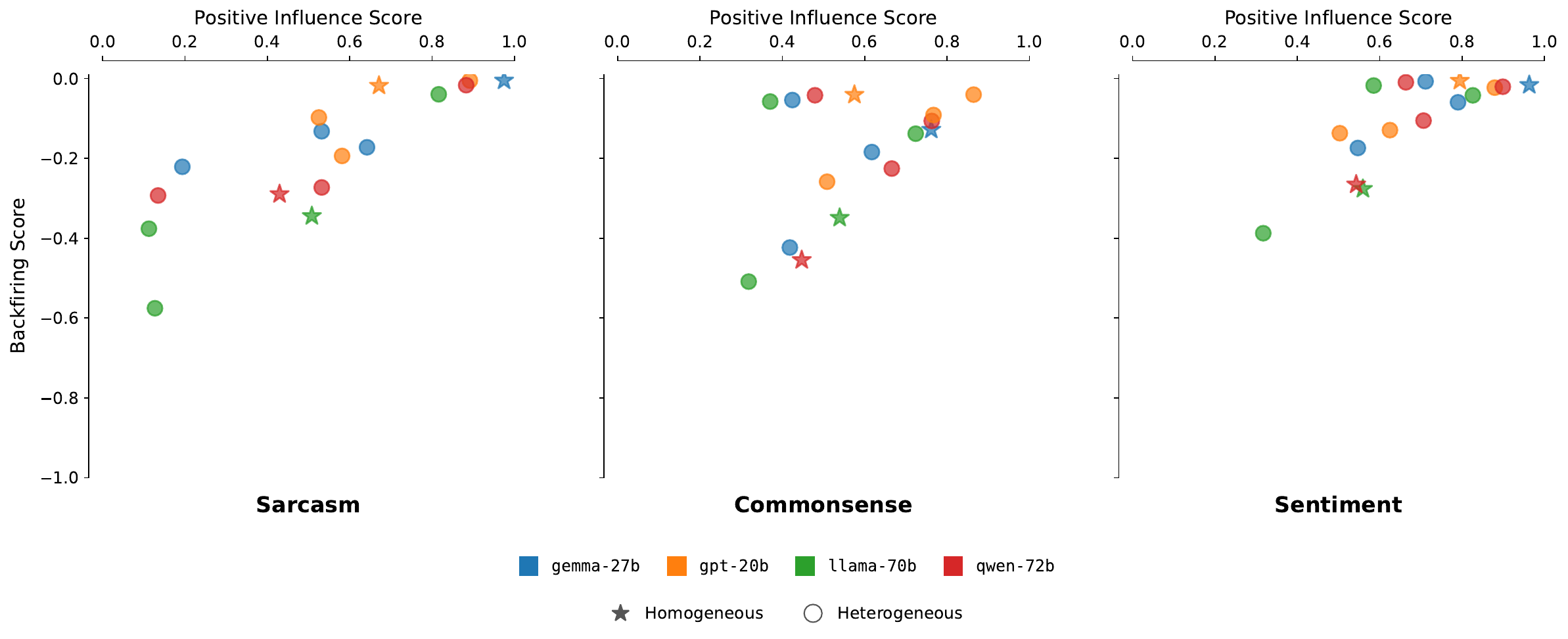}
    \caption{\textbf{Role of model heterogeneity in influence}. Model pairs plotted as a function of their positive influence score (x-axis) and backfiring score (y-axis), colored by \peer agent. Stars and circles denote homogeneous and heterogeneous systems, respectively. Non-influential \peers lie close to the origin. Those in the upper-right region are able to highly influence \judges, whereas those in the lower-left region cause strong backfiring effects.}
    \label{fig:hetero-homo-peer-based}
\end{figure*}

A detailed numerical comparison of the influence scores $\Delta_{\A|\B}$ measured in homogeneous vs. heterogeneous systems is given in Table~\ref{tab:hom-het-diff}. Figure~\ref{fig:hetero-homo-peer-based} shows each model combination as a function of its influence scores and backfiring scores when acting as peer, with an emphasis on comparing homogeneous and heterogeneous configurations.

\begin{table*}[h]
\centering
\setlength{\tabcolsep}{3pt}
\small
\begin{tabular}{l ccc ccc ccc}
\toprule
\multirow{2}{*}{\textsc{Receiver}}
& \multicolumn{3}{c}{\textsc{Sarcasm}}
& \multicolumn{3}{c}{\textsc{Commonsense}}
& \multicolumn{3}{c}{\textsc{Sentiment}} \\
\cmidrule(lr){2-4} \cmidrule(lr){5-7} \cmidrule(lr){8-10}
& \small Hom. $\Delta_{\A|\B}$ & \small Het. $\Delta_{\A|\B}$ (avg.) & \small Diff. 
& \small Hom. $\Delta_{\A|\B}$ & \small Het. $\Delta_{\A|\B}$ (avg.) & \small Diff. 
& \small Hom. $\Delta_{\A|\B}$ & \small Het. $\Delta_{\A|\B}$ (avg.) & \small Diff. \\
\midrule
\texttt{gemma-27b} & .97 & .86 & -.11 & .76 & .78 & .03  & .96 & .87 & -.09 \\
\texttt{gpt-20b}   & .67 & .14 & -.79 & .57 & .42 & -.26 & .79 & .65 & -.17 \\
\texttt{llama-70b} & .51 & .58 & -.14 & .54 & .69 & .27  & .56 & .71 &  .26 \\
\texttt{qwen-72b}  & .43 & .39 & -.09 & .45 & .42& -.07  & .54 & .46 & -.15 \\
\bottomrule
\end{tabular}

\caption{\textbf{Comparison of susceptibility to influence in homogeneous (Hom.) and heterogeneous (Het.) systems}. For each receiver, the table reports the homogeneous positive influence score, the average positive influence score across heterogeneous interactions, and the percentage difference between the two. Positive differences indicates that the model becomes more susceptible to influence when transitioning from a homogeneous system to a heterogeneous system, while negative differences indicate that the model becomes less susceptible.}
\label{tab:hom-het-diff}
\end{table*}

\clearpage

\section{Prompt Variations for Sensitivity analysis}\label{app:sensitivity-prompts}

\begin{figure*}[htbp]  
    \centering
    \includegraphics[width=1.0\textwidth]{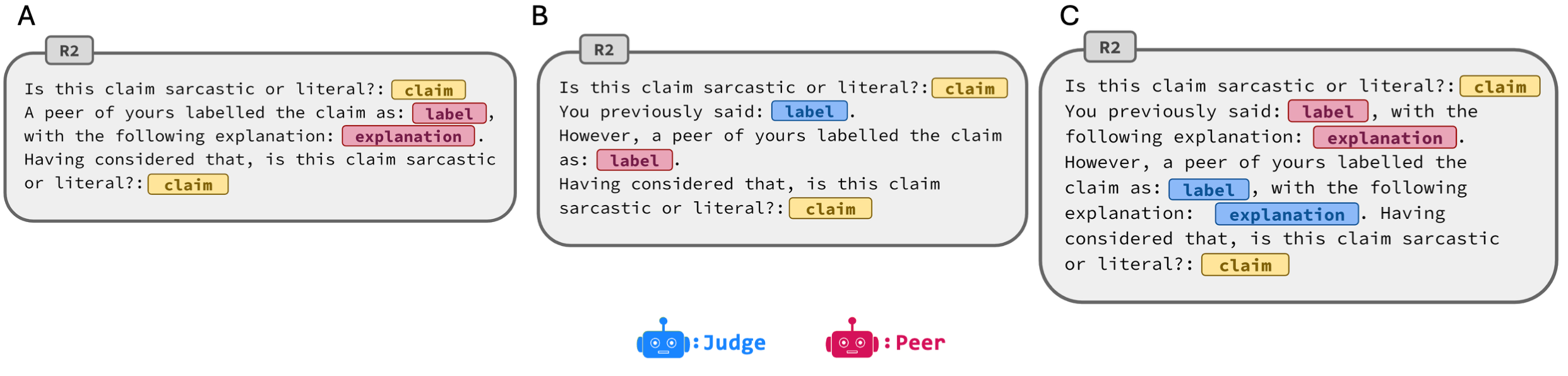}
    \caption{\textbf{Prompts for sensitivity analysis.} Prompt variations implementing a memoryless \judge (A), no explanations given (B), and swap between \judgess and \peerss label-explanation pairs (C).}
    \label{fig:prompts-robustness}
\end{figure*}

Prompt formulations used for sensitivity analysis are presented in Figure~\ref{fig:prompts-robustness}.

\section{Comparison of Model Explanations}\label{app:qual}

We examine explanation quality more closely, motivated by the swap sensitivity analysis, where the opinions of the two agents are presented from reversed roles. 
The results of the analysis suggest that \texttt{gpt-20b} consistently produces more compelling explanations than \texttt{llama-70b}. 
Regardless of whether an explanation is framed as a peer opinion or the model's own prior, both models show a systematic preference toward \texttt{gpt-20b}'s reasoning.

From a quantitative perspective, \texttt{gpt-20b} produces considerably longer explanations than \texttt{llama-70b}, with an average of $23$ words per explanation compared to $7$.
This difference in length is also reflected in the quality of the explanations each model produces. 
With longer explanations \texttt{gpt-20b} is able to construct more elaborate arguments, whereas \texttt{llama-70b} tends to produce brief statements that assert a label without providing a supporting explanation (see examples below). 
This pattern is consistent across all interaction outcomes observed in the swap analysis.
This suggests that explanation quality is a contributing factor to a peer's ability to influence a receiver.

\vspace{0.5cm}

\begin{tcolorbox}[
    title=Model explanations,
    colback=orange!15,
    colframe=brown,
    colbacktitle=orange!35,
    coltitle=black,
    fonttitle=\bfseries,
    center title,
    boxrule=0.8pt,
    arc=0pt
]
\small
\renewcommand{\arraystretch}{1.5}
\begin{tabularx}{\textwidth}{@{} X X @{}}
\textbf{\texttt{llama-70b}}: \textit{Speaker states a fact about learning something's unconstitutionality.} & \textbf{\texttt{gpt-20b}}: \textit{The sentence dismisses any benefit, highlighting absurdity. It critiques the claim’s premise, implying the speaker found the content unworthy or nonsense, typical sardonic tone.}\\[1.5em]
\textbf{\texttt{llama-70b}}: \textit{Tone implies annoyance or frustration.} & \textbf{\texttt{gpt-20b}}: \textit{The sentence is a straightforward inquiry about the availability of a release. It lacks the exaggerated tone, irony, or context typical of sarcasm.} \\[1.5em]
\textbf{\texttt{llama-70b}}: \textit{Refers to past appearance as Marshmello} & \textbf{\texttt{gpt-20b}}: \textit{The claim humorously implies someone claimed to be Marshmello, an unlikely identity, suggesting exaggeration or jest rather than fact.}
\end{tabularx}
\end{tcolorbox}

\clearpage

\section{Classification Performance} \label{app:f1_scores}

\begin{figure*}[h!]  
    \centering
    \includegraphics[width=0.8\textwidth]{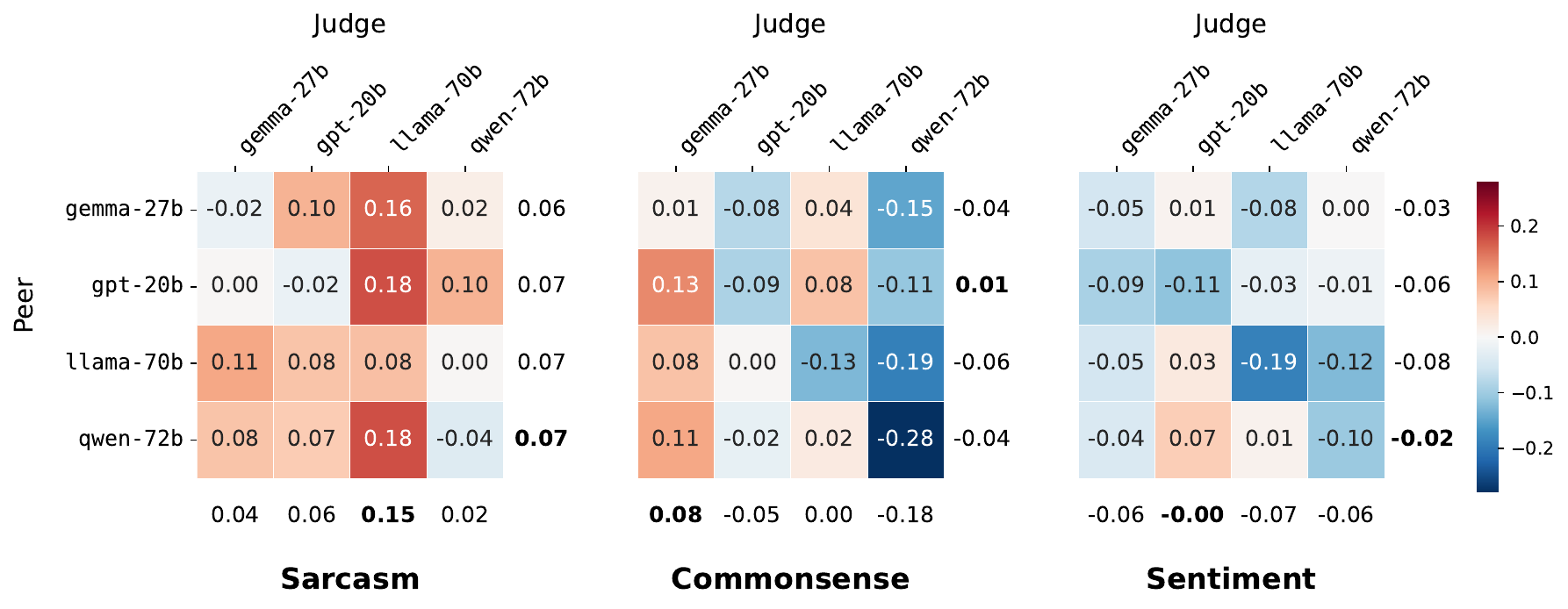}
    \caption{\textbf{Effect of peer interaction on classification performance.} Differences between F1 classification score in round 2 with the respective F1 score obtained by the Judge in isolation in round 1. Positive values indicate a performance improvement after \judge-\peer interaction.}
    \label{fig:f1_inter}
\end{figure*}

\begin{figure*}[h!]  
    \centering
    \includegraphics[width=0.7\textwidth]{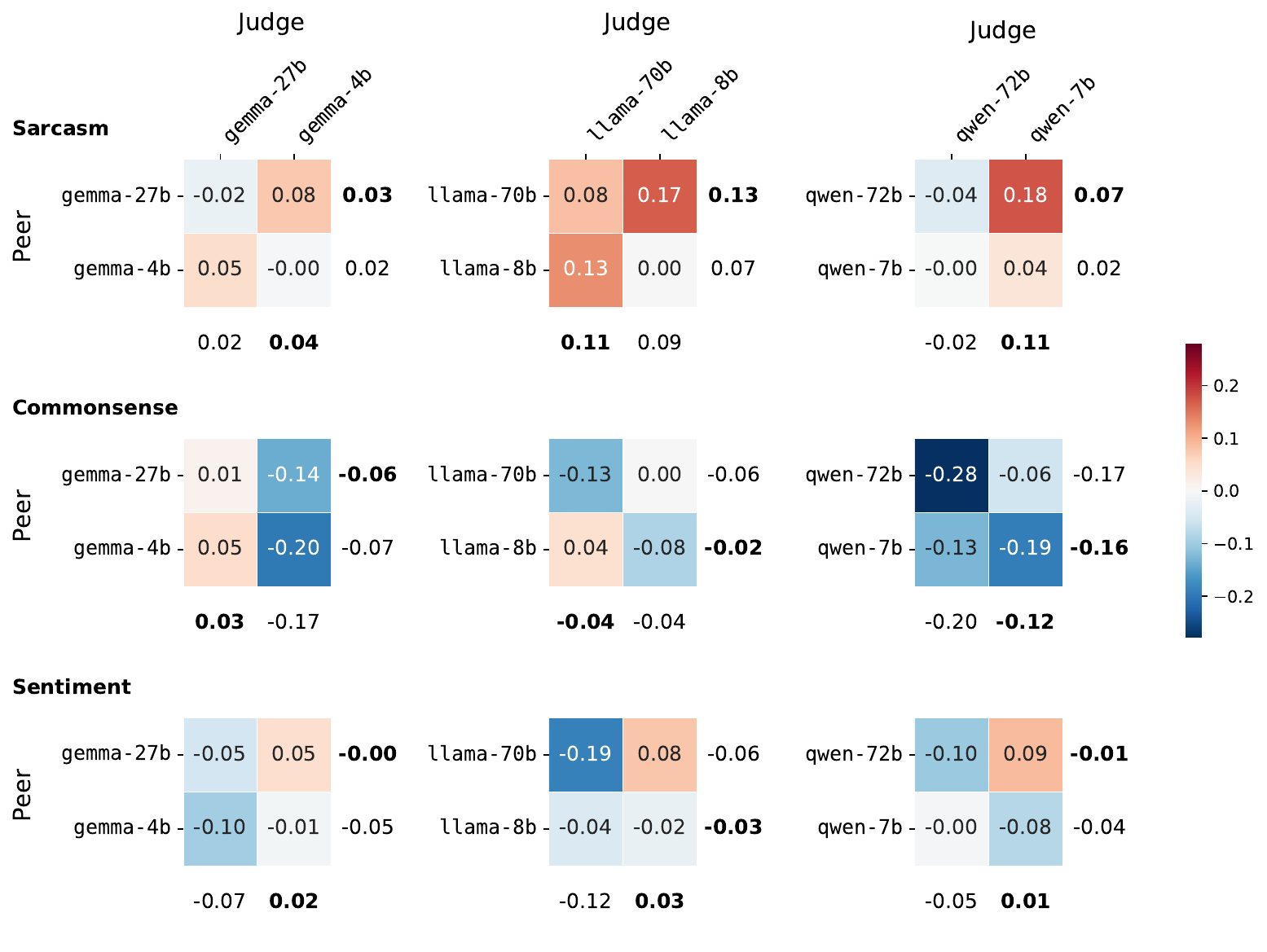}
    \caption{\textbf{Effect of peer interaction on classification performance for models of different size from the same family.} Differences between F1 classification score in round 2 with the respective F1 score obtained by the Judge in isolation in round 1, across model pairs from the same family. Positive values indicate a performance improvement after \judge-\peer interaction.}
    \label{fig:f1_intra}
\end{figure*}

Figures~\ref{fig:f1_inter} and~\ref{fig:f1_inter} show the difference of F1 classification scores between round 2, when the label is influenced by the \peer, and round 1, when the \judge performs a classification in isolation. The F1 score is the harmonic mean between precision and recall.

\clearpage

\section{Influence Scores on Match Type-Level}\label{app:match-type-scores}

We report the influence scores and backfiring scores, and count of instances they are computed from, for each match type and dataset below.

\begin{center}
\scriptsize
\begin{longtable}{l l c c r c c r c c r}
\toprule
\multirow{2}{*}{\textsc{Judge} $\boldsymbol{\leftarrow}$ \textsc{Peer}} & \multirow{2}{*}{\textsc{Match Type}} & \multicolumn{3}{c}{\textsc{Sarcasm}} & \multicolumn{3}{c}{\textsc{Commonsense}} & \multicolumn{3}{c}{\textsc{Sentiment}} \\
\cmidrule(lr){3-5} \cmidrule(lr){6-8} \cmidrule(lr){9-11} \\
 & & \tiny $\Delta_{\A|\B}$ & \tiny $\Delta^-_{\A|\B}$ & \tiny Count & \tiny $\Delta_{\A|\B}$ & \tiny $\Delta^-_{\A|\B}$ & \tiny Count & \tiny $\Delta_{\A|\B}$ & \tiny $\Delta^-_{\A|\B}$ & \tiny Count \\
\arrayrulecolor{black!80}\midrule
\endfirsthead
\toprule
\multirow{2}{*}{\textsc{Judge} $\boldsymbol{\leftarrow}$ \textsc{Peer}} & \multirow{2}{*}{\textsc{Match Type}} & \multicolumn{3}{c}{\textsc{Sarcasm}} & \multicolumn{3}{c}{\textsc{Commonsense}} & \multicolumn{3}{c}{\textsc{Sentiment}} \\
\cmidrule(lr){3-5} \cmidrule(lr){6-8} \cmidrule(lr){9-11} \\
 & & \tiny $\Delta_{\A|\B}$ & \tiny $\Delta^-_{\A|\B}$ & \tiny Count & \tiny $\Delta_{\A|\B}$ & \tiny $\Delta^-_{\A|\B}$ & \tiny Count & \tiny $\Delta_{\A|\B}$ & \tiny $\Delta^-_{\A|\B}$ & \tiny Count \\
\arrayrulecolor{black!80}\midrule
\endhead
\midrule \multicolumn{11}{r}{\tiny\itshape continued\ldots} \\
\endfoot
\bottomrule
\multicolumn{11}{c}{}\\
\caption{\textbf{Detailed influence scores ($\Delta_{\A|\B}$) and backfiring scores ($\Delta^-_{\A|\B}$).} The scores are reported together with the corresponding number of instances from which they were computed. ‘--’ indicates match types for which the score cannot be computed, as neither no positive nor negative shifts can occur given the label certainty of the \judge. Recall that $\text{0:0}$ and $\text{1:1}$ are agreeing configurations, and the remaining match types describe disagreeing configurations.}
\label{tab:large}
\endlastfoot
\texttt{gemma-27b} $\leftarrow$ \texttt{gemma-27b} & 0:0 & -- & 0.00 & 1,000 & -- & -0.00 & 1,000 & -- & 0.00 & 1,000 \\
 & 1:1 & -- & 0.00 & 1,000 & -- & -0.02 & 1,000 & -- & 0.00 & 1,000 \\
 & B:B & 0.97 & -0.01 & 5,644 & 0.76 & -0.13 & 482 & 0.96 & -0.02 & 2,668 \\
\arrayrulecolor{gray!80}\midrule
\texttt{gemma-4b} $\leftarrow$ \texttt{gemma-27b} & 0:0 & -- & -0.01 & 1,000 & -- & -0.00 & 1,000 & -- & -- & -- \\
 & 0:1 & 0.99 & -- & 1,993 & 0.15 & -- & 1,587 & 0.65 & -- & 5,200 \\
 & 0:B & 0.97 & -- & 305 & 0.09 & -- & 156 & 0.60 & -- & 846 \\
 & 1:0 & 0.76 & -- & 6,131 & 0.83 & -- & 768 & 0.81 & -- & 2,042 \\
 & 1:1 & -- & -0.00 & 1,000 & -- & -0.03 & 1,000 & -- & -- & -- \\
 & 1:B & 0.70 & -- & 2,439 & 0.87 & -- & 67 & 0.75 & -- & 419 \\
 & B:0 & 0.95 & -0.02 & 411 & 0.93 & -0.06 & 187 & 0.90 & -0.05 & 396 \\
 & B:1 & 0.99 & -0.00 & 713 & 0.22 & -0.66 & 288 & 0.86 & -0.07 & 745 \\
 & B:B & 0.97 & -0.00 & 156 & 0.38 & -0.32 & 34 & 0.85 & -0.10 & 136 \\
\arrayrulecolor{gray!80}\midrule
\texttt{gpt-20b} $\leftarrow$ \texttt{gemma-27b} & 0:0 & -- & -0.00 & 1,000 & -- & -- & -- & -- & -- & -- \\
 & 0:1 & 0.19 & -- & 7,000 & 0.19 & -- & 600 & 0.53 & -- & 952 \\
 & 0:B & 0.10 & -- & 1,857 & 0.14 & -- & 99 & 0.57 & -- & 277 \\
 & 1:0 & 0.18 & -- & 18 & 0.37 & -- & 75 & 0.71 & -- & 972 \\
 & 1:1 & -- & -0.01 & 1,000 & -- & -- & -- & -- & -- & -- \\
 & 1:B & 0.12 & -- & 15 & 0.56 & -- & 20 & 0.65 & -- & 254 \\
 & B:0 & 0.21 & -0.26 & 1,811 & 0.71 & -0.03 & 1,026 & 0.91 & -0.00 & 5,989 \\
 & B:1 & 0.34 & -0.14 & 7,000 & 0.50 & -0.06 & 2,162 & 0.81 & -0.01 & 7,000 \\
 & B:B & 0.21 & -0.27 & 1,900 & 0.51 & -0.07 & 244 & 0.80 & -0.01 & 1,606 \\
\arrayrulecolor{gray!80}\midrule
\texttt{llama-70b} $\leftarrow$ \texttt{gemma-27b} & 0:0 & -- & -0.01 & 1,000 & -- & -0.02 & 1,000 & -- & -- & -- \\
 & 0:1 & 0.87 & -- & 7,000 & 0.51 & -- & 842 & 0.66 & -- & 710 \\
 & 0:B & 0.79 & -- & 2,036 & 0.37 & -- & 126 & 0.64 & -- & 224 \\
 & 1:0 & 0.33 & -- & 421 & 0.72 & -- & 480 & 0.84 & -- & 4,394 \\
 & 1:1 & -- & -0.03 & 1,000 & -- & -0.04 & 1,000 & -- & -- & -- \\
 & 1:B & 0.35 & -- & 398 & 0.77 & -- & 100 & 0.77 & -- & 1,033 \\
 & B:0 & 0.47 & -0.32 & 524 & 0.85 & -0.09 & 114 & 0.95 & -0.02 & 970 \\
 & B:1 & 0.96 & -0.01 & 7,000 & 0.57 & -0.33 & 188 & 0.77 & -0.12 & 273 \\
 & B:B & 0.72 & -0.19 & 776 & 0.54 & -0.14 & 30 & 0.91 & -0.04 & 154 \\
\arrayrulecolor{gray!80}\midrule
\texttt{qwen-72b} $\leftarrow$ \texttt{gemma-27b} & 0:0 & -- & -0.00 & 1,000 & -- & -0.00 & 1,000 & -- & -- & -- \\
 & 0:1 & 0.40 & -- & 7,000 & 0.11 & -- & 1,001 & 0.41 & -- & 2,097 \\
 & 0:B & 0.26 & -- & 1,547 & 0.12 & -- & 146 & 0.34 & -- & 592 \\
 & 1:0 & 0.60 & -- & 1,123 & 0.59 & -- & 379 & 0.64 & -- & 1,723 \\
 & 1:1 & -- & -0.00 & 1,000 & -- & -0.03 & 1,000 & -- & -- & -- \\
 & 1:B & 0.56 & -- & 881 & 0.63 & -- & 72 & 0.49 & -- & 589 \\
 & B:0 & 0.76 & -0.06 & 749 & 0.78 & -0.12 & 134 & 0.74 & -0.12 & 654 \\
 & B:1 & 0.57 & -0.16 & 5,797 & 0.19 & -0.63 & 285 & 0.61 & -0.20 & 788 \\
 & B:B & 0.58 & -0.18 & 788 & 0.51 & -0.52 & 46 & 0.60 & -0.21 & 306 \\
\arrayrulecolor{gray!80}\midrule
\texttt{gemma-27b} $\leftarrow$ \texttt{gemma-4b} & 0:0 & -- & 0.00 & 1,000 & -- & -0.01 & 1,000 & -- & -- & -- \\
 & 0:1 & 0.90 & -- & 6,131 & 0.38 & -- & 768 & 0.58 & -- & 2,042 \\
 & 0:B & 0.84 & -- & 411 & 0.31 & -- & 187 & 0.58 & -- & 396 \\
 & 1:0 & 0.59 & -- & 1,993 & 0.51 & -- & 1,587 & 0.89 & -- & 5,200 \\
 & 1:1 & -- & -0.00 & 1,000 & -- & -0.02 & 1,000 & -- & -- & -- \\
 & 1:B & 0.57 & -- & 713 & 0.48 & -- & 288 & 0.81 & -- & 745 \\
 & B:0 & 0.88 & -0.05 & 305 & 0.71 & -0.15 & 156 & 0.98 & -0.00 & 846 \\
 & B:1 & 0.95 & -0.01 & 2,439 & 0.60 & -0.37 & 67 & 0.88 & -0.07 & 419 \\
 & B:B & 0.92 & -0.01 & 156 & 0.56 & -0.24 & 34 & 0.97 & -0.03 & 136 \\
\arrayrulecolor{gray!80}\midrule
\texttt{gemma-4b} $\leftarrow$ \texttt{gemma-4b} & 0:0 & -- & -0.03 & 1,000 & -- & -0.00 & 1,000 & -- & 0.00 & 1,000 \\
 & 1:1 & -- & -0.00 & 1,000 & -- & -0.03 & 1,000 & -- & -0.00 & 1,000 \\
 & B:B & 0.90 & -0.04 & 2,404 & 0.40 & -0.48 & 984 & 0.63 & -0.22 & 2,418 \\
\arrayrulecolor{gray!80}\midrule
\texttt{gemma-27b} $\leftarrow$ \texttt{gpt-20b} & 0:0 & -- & 0.00 & 1,000 & -- & -0.00 & 1,000 & -- & -- & -- \\
 & 0:1 & 0.97 & -- & 18 & 0.94 & -- & 75 & 0.75 & -- & 972 \\
 & 0:B & 0.98 & -- & 1,811 & 0.76 & -- & 1,026 & 0.67 & -- & 5,989 \\
 & 1:0 & 0.83 & -- & 7,000 & 0.92 & -- & 600 & 0.96 & -- & 952 \\
 & 1:1 & -- & 0.00 & 1,000 & -- & -0.00 & 1,000 & -- & -- & -- \\
 & 1:B & 0.54 & -- & 7,000 & 0.73 & -- & 2,162 & 0.89 & -- & 7,000 \\
 & B:0 & 0.97 & -0.00 & 1,857 & 0.94 & -0.02 & 99 & 1.00 & 0.00 & 277 \\
 & B:1 & 0.99 & 0.00 & 15 & 0.94 & 0.00 & 20 & 0.93 & -0.05 & 254 \\
 & B:B & 0.96 & -0.01 & 1,900 & 0.83 & -0.10 & 244 & 0.95 & -0.02 & 1,606 \\
\arrayrulecolor{gray!80}\midrule
\texttt{gpt-20b} $\leftarrow$ \texttt{gpt-20b} & 0:0 & -- & -0.01 & 1,000 & -- & -- & -- & -- & -- & -- \\
 & 1:1 & -- & -0.01 & 1,000 & -- & -- & -- & -- & -- & -- \\
 & B:B & 0.67 & -0.02 & 14,000 & 0.57 & -0.04 & 6,620 & 0.79 & -0.01 & 14,000 \\
\arrayrulecolor{gray!80}\midrule
\texttt{llama-70b} $\leftarrow$ \texttt{gpt-20b} & 0:0 & -- & -0.02 & 1,000 & -- & -0.00 & 1,000 & -- & -- & -- \\
 & 0:1 & 0.90 & -- & 267 & 0.85 & -- & 97 & 0.45 & -- & 548 \\
 & 0:B & 0.86 & -- & 7,000 & 0.59 & -- & 1,210 & 0.40 & -- & 3,736 \\
 & 1:0 & 0.25 & -- & 7,000 & 0.86 & -- & 415 & 0.72 & -- & 2,393 \\
 & 1:1 & -- & -0.00 & 1,000 & -- & -0.00 & 1,000 & -- & -- & -- \\
 & 1:B & 0.11 & -- & 7,000 & 0.67 & -- & 1,914 & 0.48 & -- & 7,000 \\
 & B:0 & 0.37 & -0.35 & 3,742 & 0.89 & -0.04 & 111 & 0.90 & -0.04 & 596 \\
 & B:1 & 0.97 & -0.00 & 345 & 0.79 & -0.06 & 22 & 0.61 & -0.26 & 139 \\
 & B:B & 0.61 & -0.22 & 12,432 & 0.72 & -0.18 & 368 & 0.81 & -0.09 & 1,170 \\
\arrayrulecolor{gray!80}\midrule
\texttt{qwen-72b} $\leftarrow$ \texttt{gpt-20b} & 0:0 & -- & 0.00 & 1,000 & -- & -0.00 & 1,000 & -- & -- & -- \\
 & 0:1 & 0.63 & -- & 81 & 0.38 & -- & 76 & 0.35 & -- & 1,006 \\
 & 0:B & 0.43 & -- & 5,128 & 0.16 & -- & 1,327 & 0.24 & -- & 6,151 \\
 & 1:0 & 0.39 & -- & 7,000 & 0.81 & -- & 245 & 0.60 & -- & 752 \\
 & 1:1 & -- & 0.00 & 1,000 & -- & -0.00 & 1,000 & -- & -- & -- \\
 & 1:B & 0.14 & -- & 7,000 & 0.55 & -- & 1,710 & 0.33 & -- & 7,000 \\
 & B:0 & 0.70 & -0.07 & 3,305 & 0.91 & -0.03 & 123 & 0.77 & -0.06 & 361 \\
 & B:1 & 0.79 & -0.07 & 110 & 0.26 & -0.37 & 47 & 0.62 & -0.18 & 291 \\
 & B:B & 0.59 & -0.15 & 7,050 & 0.49 & -0.37 & 544 & 0.61 & -0.17 & 1,886 \\
\arrayrulecolor{gray!80}\midrule
\texttt{gemma-27b} $\leftarrow$ \texttt{llama-70b} & 0:0 & -- & 0.00 & 1,000 & -- & -0.00 & 1,000 & -- & -- & -- \\
 & 0:1 & 0.94 & -- & 421 & 0.72 & -- & 480 & 0.67 & -- & 4,394 \\
 & 0:B & 0.92 & -- & 524 & 0.58 & -- & 114 & 0.56 & -- & 970 \\
 & 1:0 & 0.64 & -- & 7,000 & 0.75 & -- & 842 & 0.94 & -- & 710 \\
 & 1:1 & -- & -0.00 & 1,000 & -- & -0.01 & 1,000 & -- & -- & -- \\
 & 1:B & 0.51 & -- & 7,000 & 0.71 & -- & 188 & 0.93 & -- & 273 \\
 & B:0 & 0.85 & -0.06 & 2,036 & 0.85 & -0.10 & 126 & 0.99 & -0.00 & 224 \\
 & B:1 & 0.95 & -0.01 & 398 & 0.72 & -0.15 & 100 & 0.83 & -0.08 & 1,033 \\
 & B:B & 0.89 & -0.05 & 776 & 0.73 & -0.17 & 30 & 0.88 & -0.04 & 154 \\
\arrayrulecolor{gray!80}\midrule
\texttt{gpt-20b} $\leftarrow$ \texttt{llama-70b} & 0:0 & -- & -0.01 & 1,000 & -- & -- & -- & -- & -- & -- \\
 & 0:1 & 0.14 & -- & 7,000 & 0.18 & -- & 415 & 0.39 & -- & 2,393 \\
 & 0:B & 0.12 & -- & 3,742 & 0.11 & -- & 111 & 0.38 & -- & 596 \\
 & 1:0 & 0.04 & -- & 267 & 0.37 & -- & 97 & 0.60 & -- & 548 \\
 & 1:1 & -- & -0.02 & 1,000 & -- & -- & -- & -- & -- & -- \\
 & 1:B & 0.04 & -- & 345 & 0.42 & -- & 22 & 0.56 & -- & 139 \\
 & B:0 & 0.06 & -0.48 & 7,000 & 0.61 & -0.04 & 1,210 & 0.85 & -0.01 & 3,736 \\
 & B:1 & 0.25 & -0.21 & 7,000 & 0.46 & -0.06 & 1,914 & 0.65 & -0.02 & 7,000 \\
 & B:B & 0.14 & -0.44 & 12,432 & 0.45 & -0.08 & 368 & 0.68 & -0.02 & 1,170 \\
\arrayrulecolor{gray!80}\midrule
\texttt{llama-70b} $\leftarrow$ \texttt{llama-70b} & 0:0 & -- & -0.08 & 1,000 & -- & -0.02 & 1,000 & -- & -0.00 & 1,000 \\
 & 1:1 & -- & -0.05 & 1,000 & -- & -0.05 & 1,000 & -- & -0.01 & 1,000 \\
 & B:B & 0.51 & -0.34 & 14,000 & 0.54 & -0.35 & 634 & 0.56 & -0.28 & 2,640 \\
\arrayrulecolor{gray!80}\midrule
\texttt{llama-8b} $\leftarrow$ \texttt{llama-70b} & 0:0 & -- & -0.83 & 369 & -- & -0.42 & 1,000 & -- & -0.00 & 1,000 \\
 & 0:1 & 0.84 & -- & 20 & 0.59 & -- & 77 & 0.31 & -- & 1,095 \\
 & 0:B & 0.86 & -- & 7 & 0.47 & -- & 13 & 0.31 & -- & 331 \\
 & 1:0 & 0.00 & -- & 7,000 & 0.66 & -- & 1,071 & 0.26 & -- & 1,697 \\
 & 1:1 & -- & -0.02 & 1,000 & -- & -0.20 & 1,000 & -- & -0.00 & 1,000 \\
 & 1:B & 0.00 & -- & 7,000 & 0.60 & -- & 152 & 0.28 & -- & 415 \\
 & B:0 & 0.00 & -0.91 & 7,000 & 0.63 & -0.07 & 2,731 & 0.20 & -0.36 & 4,044 \\
 & B:1 & 0.93 & -0.00 & 6,326 & 0.36 & -0.25 & 1,126 & 0.25 & -0.31 & 3,630 \\
 & B:B & 0.25 & -0.27 & 4,748 & 0.48 & -0.22 & 292 & 0.20 & -0.36 & 1,148 \\
\arrayrulecolor{gray!80}\midrule
\texttt{qwen-72b} $\leftarrow$ \texttt{llama-70b} & 0:0 & -- & -0.00 & 1,000 & -- & -0.00 & 1,000 & -- & -- & -- \\
 & 0:1 & 0.10 & -- & 2,404 & 0.08 & -- & 715 & 0.11 & -- & 4,578 \\
 & 0:B & 0.07 & -- & 1,738 & 0.04 & -- & 174 & 0.05 & -- & 968 \\
 & 1:0 & 0.13 & -- & 7,000 & 0.46 & -- & 465 & 0.45 & -- & 588 \\
 & 1:1 & -- & 0.00 & 1,000 & -- & -0.02 & 1,000 & -- & -- & -- \\
 & 1:B & 0.08 & -- & 7,000 & 0.50 & -- & 113 & 0.40 & -- & 249 \\
 & B:0 & 0.23 & -0.47 & 3,266 & 0.67 & -0.22 & 167 & 0.67 & -0.17 & 216 \\
 & B:1 & 0.14 & -0.63 & 2,471 & 0.14 & -0.79 & 242 & 0.23 & -0.56 & 1,275 \\
 & B:B & 0.14 & -0.63 & 2,406 & 0.34 & -0.52 & 60 & 0.31 & -0.43 & 206 \\
\arrayrulecolor{gray!80}\midrule
\texttt{llama-70b} $\leftarrow$ \texttt{llama-8b} & 0:0 & -- & -0.02 & 369 & -- & -0.01 & 1,000 & -- & -0.00 & 1,000 \\
 & 0:1 & 0.67 & -- & 7,000 & 0.23 & -- & 1,071 & 0.33 & -- & 1,697 \\
 & 0:B & 0.59 & -- & 7,000 & 0.15 & -- & 2,731 & 0.20 & -- & 4,044 \\
 & 1:0 & 0.15 & -- & 20 & 0.47 & -- & 77 & 0.58 & -- & 1,095 \\
 & 1:1 & -- & -0.03 & 1,000 & -- & -0.05 & 1,000 & -- & -0.01 & 1,000 \\
 & 1:B & 0.09 & -- & 6,326 & 0.53 & -- & 1,126 & 0.48 & -- & 3,630 \\
 & B:0 & 0.33 & -0.44 & 7 & 0.58 & -0.36 & 13 & 0.77 & -0.10 & 331 \\
 & B:1 & 0.89 & -0.03 & 7,000 & 0.33 & -0.43 & 152 & 0.46 & -0.32 & 415 \\
 & B:B & 0.57 & -0.32 & 4,748 & 0.48 & -0.33 & 292 & 0.57 & -0.24 & 1,148 \\
\arrayrulecolor{gray!80}\midrule
\texttt{llama-8b} $\leftarrow$ \texttt{llama-8b} & 0:0 & -- & -0.68 & 396 & -- & -0.30 & 1,000 & -- & -0.00 & 1,000 \\
 & 1:1 & -- & -0.01 & 1,000 & -- & -0.17 & 1,000 & -- & -0.00 & 1,000 \\
 & B:B & 0.26 & -0.28 & 14,000 & 0.50 & -0.20 & 8,014 & 0.29 & -0.28 & 14,000 \\
\arrayrulecolor{gray!80}\midrule
\texttt{gemma-27b} $\leftarrow$ \texttt{qwen-72b} & 0:0 & -- & 0.00 & 1,000 & -- & -0.00 & 1,000 & -- & -- & -- \\
 & 0:1 & 0.98 & -- & 1,123 & 0.77 & -- & 379 & 0.78 & -- & 1,723 \\
 & 0:B & 0.95 & -- & 749 & 0.61 & -- & 134 & 0.75 & -- & 654 \\
 & 1:0 & 0.76 & -- & 7,000 & 0.82 & -- & 1,001 & 0.95 & -- & 2,097 \\
 & 1:1 & -- & 0.00 & 1,000 & -- & -0.01 & 1,000 & -- & -- & -- \\
 & 1:B & 0.65 & -- & 5,797 & 0.70 & -- & 285 & 0.94 & -- & 788 \\
 & B:0 & 0.92 & -0.03 & 1,547 & 0.81 & -0.07 & 146 & 0.99 & -0.00 & 592 \\
 & B:1 & 0.98 & -0.00 & 881 & 0.78 & -0.12 & 72 & 0.92 & -0.05 & 589 \\
 & B:B & 0.93 & -0.02 & 788 & 0.83 & -0.13 & 46 & 0.97 & -0.01 & 306 \\
\arrayrulecolor{gray!80}\midrule
\texttt{gpt-20b} $\leftarrow$ \texttt{qwen-72b} & 0:0 & -- & -0.00 & 1,000 & -- & -- & -- & -- & -- & -- \\
 & 0:1 & 0.14 & -- & 7,000 & 0.27 & -- & 245 & 0.53 & -- & 752 \\
 & 0:B & 0.10 & -- & 3,305 & 0.20 & -- & 123 & 0.55 & -- & 361 \\
 & 1:0 & 0.07 & -- & 81 & 0.53 & -- & 76 & 0.59 & -- & 1,006 \\
 & 1:1 & -- & -0.02 & 1,000 & -- & -- & -- & -- & -- & -- \\
 & 1:B & 0.07 & -- & 110 & 0.49 & -- & 47 & 0.53 & -- & 291 \\
 & B:0 & 0.13 & -0.34 & 5,128 & 0.74 & -0.02 & 1,327 & 0.85 & -0.01 & 6,151 \\
 & B:1 & 0.27 & -0.18 & 7,000 & 0.58 & -0.04 & 1,710 & 0.81 & -0.01 & 7,000 \\
 & B:B & 0.16 & -0.35 & 7,050 & 0.54 & -0.07 & 544 & 0.79 & -0.02 & 1,886 \\
\arrayrulecolor{gray!80}\midrule
\texttt{llama-70b} $\leftarrow$ \texttt{qwen-72b} & 0:0 & -- & -0.01 & 1,000 & -- & -0.01 & 1,000 & -- & -- & -- \\
 & 0:1 & 0.80 & -- & 7,000 & 0.63 & -- & 465 & 0.61 & -- & 588 \\
 & 0:B & 0.73 & -- & 3,266 & 0.47 & -- & 167 & 0.53 & -- & 216 \\
 & 1:0 & 0.20 & -- & 2,404 & 0.78 & -- & 715 & 0.73 & -- & 4,578 \\
 & 1:1 & -- & -0.03 & 1,000 & -- & -0.02 & 1,000 & -- & -- & -- \\
 & 1:B & 0.17 & -- & 2,471 & 0.65 & -- & 242 & 0.62 & -- & 1,275 \\
 & B:0 & 0.31 & -0.50 & 1,738 & 0.85 & -0.09 & 174 & 0.91 & -0.04 & 968 \\
 & B:1 & 0.92 & -0.03 & 7,000 & 0.58 & -0.30 & 113 & 0.73 & -0.18 & 249 \\
 & B:B & 0.59 & -0.30 & 2,406 & 0.70 & -0.29 & 60 & 0.82 & -0.09 & 206 \\
\arrayrulecolor{gray!80}\midrule
\texttt{qwen-72b} $\leftarrow$ \texttt{qwen-72b} & 0:0 & -- & -0.00 & 1,000 & -- & -0.00 & 1,000 & -- & 0.00 & 1,000 \\
 & 1:1 & -- & -0.00 & 1,000 & -- & -0.02 & 1,000 & -- & 0.00 & 1,000 \\
 & B:B & 0.43 & -0.29 & 13,880 & 0.45 & -0.45 & 884 & 0.54 & -0.27 & 3,190 \\
\arrayrulecolor{gray!80}\midrule
\texttt{qwen-7b} $\leftarrow$ \texttt{qwen-72b} & 0:0 & -- & -0.00 & 1,000 & -- & -0.00 & 1,000 & -- & -- & -- \\
 & 0:1 & 0.75 & -- & 128 & 0.26 & -- & 1,830 & 0.56 & -- & 1,371 \\
 & 0:B & 0.77 & -- & 95 & 0.18 & -- & 372 & 0.52 & -- & 528 \\
 & 1:0 & 0.19 & -- & 7,000 & 0.89 & -- & 168 & 0.55 & -- & 3,472 \\
 & 1:1 & -- & -0.00 & 1,000 & -- & -0.02 & 1,000 & -- & -- & -- \\
 & 1:B & 0.11 & -- & 6,790 & 0.80 & -- & 53 & 0.48 & -- & 911 \\
 & B:0 & 0.28 & -0.42 & 465 & 0.84 & -0.13 & 75 & 0.62 & -0.17 & 993 \\
 & B:1 & 0.78 & -0.12 & 85 & 0.31 & -0.50 & 203 & 0.61 & -0.22 & 575 \\
 & B:B & 0.52 & -0.27 & 110 & 0.57 & -0.24 & 32 & 0.49 & -0.24 & 310 \\
\arrayrulecolor{gray!80}\midrule
\texttt{qwen-72b} $\leftarrow$ \texttt{qwen-7b} & 0:0 & -- & -0.00 & 1,000 & -- & -0.00 & 1,000 & -- & -- & -- \\
 & 0:1 & 0.11 & -- & 7,000 & 0.06 & -- & 168 & 0.16 & -- & 3,472 \\
 & 0:B & 0.05 & -- & 465 & 0.05 & -- & 75 & 0.13 & -- & 993 \\
 & 1:0 & 0.18 & -- & 128 & 0.32 & -- & 1,830 & 0.35 & -- & 1,371 \\
 & 1:1 & -- & -0.00 & 1,000 & -- & -0.02 & 1,000 & -- & -- & -- \\
 & 1:B & 0.15 & -- & 85 & 0.26 & -- & 203 & 0.24 & -- & 575 \\
 & B:0 & 0.27 & -0.37 & 95 & 0.51 & -0.29 & 372 & 0.50 & -0.28 & 528 \\
 & B:1 & 0.18 & -0.50 & 6,790 & 0.11 & -0.85 & 53 & 0.35 & -0.42 & 911 \\
 & B:B & 0.21 & -0.55 & 110 & 0.34 & -0.56 & 32 & 0.34 & -0.44 & 310 \\
\arrayrulecolor{gray!80}\midrule
\texttt{qwen-7b} $\leftarrow$ \texttt{qwen-7b} & 0:0 & -- & -0.00 & 1,000 & -- & -0.02 & 1,000 & -- & -0.00 & 1,000 \\
 & 1:1 & -- & -0.00 & 1,000 & -- & -0.03 & 1,000 & -- & -0.00 & 1,000 \\
 & B:B & 0.42 & -0.33 & 1,210 & 0.40 & -0.40 & 588 & 0.47 & -0.29 & 3,446 \\
\end{longtable}
\end{center}